\documentclass[a4paper,11pt]{article}
\usepackage{amssymb,amsmath}
\usepackage{array}
\usepackage{xcolor}
\usepackage{colortbl}
\usepackage{hyperref}
\usepackage{subfigure}
\usepackage{listings}
\usepackage{comment}
\usepackage{float}
\usepackage{authblk}

\usepackage{graphicx}

\usepackage[margin=0.85in]{geometry}

\newcommand{\code}[1]{\lstinline{#1}}

\begin{document}


\title{KrakN: Transfer Learning framework for thin crack detection in infrastructure maintenance}

\author{Mateusz \.Zarski}
\author{Bartosz Wójcik}
\affil{Faculty of Civil Engineering, Silesian University of Technology, 
Akademicka 5, 44-100 Gliwice, Poland}
\affil{Chung-Ang University, School of Architecture and Building 
Science, 84 Heukseok-ro, Dongjak-gu, 06974, Seoul,
Korea}
\author{Jaros{\l}aw Adam Miszczak}
\affil{Institute of Theoretical and Applied Informatics, Polish Academy 
of Sciences, Ba{l}tycka 5, 44-100 Gliwice, Poland}

\maketitle

\begin{abstract}
Monitoring the technical condition of infrastructure is a crucial element to its maintenance. Currently applied methods are outdated, labour-intensive and inaccurate. At the same time, the latest methods using Artificial Intelligence techniques are severely limited in their application due to two main factors --- labour-intensive gathering of new datasets and high demand for computing power. We propose to utilize custom made framework --- KrakN, to overcome these limiting factors. It enables development of unique infrastructure defects detectors on digital images, achieving the accuracy of above 90\%. The framework supports semi-automatic creation of new datasets and has modest computing power requirements. It is implemented in the form of a ready-to-use software package openly distributed to the public. Thus, it can be used to immediately implement the methods proposed in this paper in the process of infrastructure management by government units, regardless of their financial capabilities.
\end{abstract}


\section{Introduction}
\label{sed:intro}

The most important features of bridge infrastructure are its operational safety and the ability to preserve it during its life cycle. These features directly affect the safety of infrastructure users. However, with the general trend of the average age of bridge structures in the world constantly increasing and exceeding 50 years for almost half of the bridges in use \cite{RoadsJapan,ASCE}, ensuring the safety of infrastructure using existing, labour-intensive methods becomes gradually more difficult. Despite the decreasing expenditure on the construction of new infrastructure members in favour of the maintenance of the existing ones \cite{Kane2019,Wso2015}, the methods currently used to ensure their safety are still largely based on an outdated assessment system of scores arbitrarily assigned to a structure by the Bridge Inspector. Such methods are error-prone and proved to be very inaccurate in real-world scenarios. Moreover, despite numerous attempts to create autonomous systems with the use of advanced computational methods to assess the technical condition of infrastructure facilities, as well as attempts to transfer the burden of technical assessment of the infrastructure to the structural health monitoring (SHM) systems built into the facility, visual assessment methods used by the Bridge Inspectors are still the basis of assessing the technical condition of infrastructure elements worldwide.

Currently, the typical way of work of the Bridge Inspector is repetitive, labour-intensive \cite{Guidelines} and prone to errors. Over a half of final bridge assessments are most likely not correct and as few as 2 to 40\% of small defects, such as like cracks, are documented at all during the inspections \cite{Phares2004, Graybeal2002}. It should be stressed out that this is usually the only level of direct contact of the infrastructure management unit with the structure and errors made at this stage are transferred to the entire process of infrastructure maintenance. . This susceptibility to human error during inspections may lead to infrastructure disasters \cite{Genova,Mumbai}. Thus, it is crucial to minimize such errors as they may cost dozens of lives and ultimately lead to undermining social confidence in infrastructure safety.

This problem has been referred to many times in the past, but despite the proposals of involvement of modern computer methods for supporting the work of the Bridge Inspector, none of the methods has been widely adopted. One of the problems is relatively low accuracy of the proposed methods in detecting structural defects in the case of using traditional image-processing techniques like edge detection and thresholding as compared to the methods utilizing Machine Learning \cite{Dorafshan2018} or high complexity and relative novelty of Deep Learning methods. This results in a narrow range of scientific units related to Civil Engineering involved in its development. Another limitation is its high demand for computing power \cite{Rhu2016} needed for algorithm training from scratch. This excludes utilization of such techniques by the local government management units operating on a limited budget and without access to appropriate computing skills.

Some of the most promising methods proposed in the past include SHM systems based on vibration measurements or structure dynamic response \cite{Shang2020,Fan2020}, image-processing color-based methods \cite{Lee2006} or utilizing various edge detection algorithms \cite{Adhikari2014,Adhikari2012,Wang2018,Li2014,Koch2014}. Systems for small-size defects detecting using stereoscopic cameras \cite{Yang2015,Yang2018} or terrestrial laser scans \cite{Rabah2013} and systems based on Convolutional Neural Networks \cite{Oh2020, Atha2018,He2020} were also extensively researched. Other approaches utilizing Machine Learning include using \textit{Fully Convolutional Networks} as in \cite{Mei2020, Dung2019, Yang2018a, Li2019} for semantic image segmentation, combined classifying-segmenting methods \cite{Ni2019} or LSTM recurrent neural networks \cite{Zhang2019}. However, their effectiveness comes with much higher computing power demand and thus lower adaptability. While most of the proposed solutions proved to be highly accurate and ready to be implemented on a large scale, they have not yet been implemented by infrastructure maintenance units at the level of local Bridge Inspectors or even at a well-financed governmental level. In contrast, the solution proposed in this paper has been designed to be used directly on the local levels of infrastructure maintenance and to provide reliable support for Bridge Inspectors.

Limitations of the methods mentioned above can be addressed through the use of the technique of Transfer Learning \cite{rosebrock2017deep}. It simultaneously gives the benefits of the accuracy of using Deep Learning methods, \emph{e.g.} in the cases of varying degrees of lighting \cite{Cha2017}, while not requiring as much training data. It also significantly reduces the need for computing power and does not require the development of an Artificial Neural Network architecture from scratch, enabling the use of models pre-trained on large, accurately labelled datasets. In addition, the methods proposed in the article can be used not only for detecting single type of defect, but they also give the possibility of building separate classifiers to detect construction or localization-specific defects by Bridge Inspectors with limited knowledge or skills in programming, that would suit their own area of work.

The main goal of the presented paper is to deliver an easy to deploy, Deep Learning based framework that can be used directly by the Bridge Inspector for detecting thin cracks on digital photos as well as for building own defect classifiers with datasets acquired during the inspections. The proposed methods deliver the accuracy of defect detection greatly surpassing the accuracy obtained in the manual approach. The algorithms and workflow of the framework are implemented in the form of a software package which provides the ability to modify its components for managing and fine-tuning classifiers according to the specific use cases. The presented software, including a sample features extracted from the dataset for thin cracks detection classifier and the pre-compiled executable files, as well as the classifier itself, is distributed under open source license and publicly accessible~\cite{KrakN}. Public availability of the software and the datasets developed for the purpose of infrastructure maintenance is crucial for deploying modern methods in this area. This issue has already been addressed in~\cite{Liu2019} where a public benchmark for crack detection system was provided~\cite{DeepCrackDataset,DeepSegmentor}. We believe that KrakN software is also an important step in popularizing the applications of artificial intelligence in infrastructure maintenance.

The rest of this paper is organized as follows. In Section~\ref{sec:methdology}, we introduce crucial elements and main building blocks of the proposed framework. Section~\ref{sec:implementation} provides implementation details, including semi-automatic building methods of the datasets. In Section \ref{sec:framework-benchmark} results of the experimental study conducted with different surface materials and on data acquired with different devices in diverse conditions are described. To highlight the advantages of the proposed approach, in Section~\ref{sec:comparative-study} we test our framework against the existing approaches and demonstrate its utilization in the multi-stage approach. Section~\ref{sec:conclusions} contains a discussion of the presented results and some concluding remarks.

\section{Principles of the proposed framework}\label{sec:methdology}

This section introduces techniques utilized to develop the framework described in this work. We stress the underlying principles of using Convolutional Neutral Networks which make them an appropriate tool for employing Transfer Learning techniques in visual object detection. We also highlight the benefits of utilizing Transfer Learning from the perspective of deployment in practical situations. A detailed description of the introduced concepts is provided in Section~\ref{sec:implementation} along with the information about the software package~\cite{KrakN}.

\subsection{Sliding window method}

In recent years, a substantial increase of interest in the subject of digital image object detection can be observed, which resulted in the creation of new algorithms, such as \textit{Fast and Faster R-CNN}, \textit{Single Shot Detectors (SSD)} or \textit{You Only Look Once (YOLO)} algorithms. Their goal was to maximize the efficiency of the algorithm in the context of the speed of its operation. Some of them have already found application in Civil Engineering as damage detectors, for example for detecting corrosion of structural steel or pavement defects~\cite{Cha2018,Du2020}. However, despite the advantages resulting from the effectiveness of their low operation latency, these algorithms have a number of drawbacks excluding them from effective and versatile use as working tools for the Bridge Inspector.

In particular, in order to achieve high performance, the above-mentioned algorithms sacrifice the number of possible objects detected in the image and are by default used on low-resolution images \cite{Du2020}. However, the most important advantage of the object detection algorithms -- their speed of operation -- is not critical in the image analysis for infrastructure inspection, as this process does not need to be carried out during obtaining of the inspection photos. Furthermore, in order to detect small defects such as thin cracks of width under 0.2 mm, a certain resolution (high image pixel density) has to be maintained. Another important premise against the use of object detectors is their relatively high complexity in the process of preparing and labelling training data.

For these reasons, for the purpose of the present work it was decided to use the sliding window method as in \cite{Cha2018,Yokoyama2017}. This approach resolves the problem of the input image resolution and thus enables the detection of think cracks. Moreover, it enables the use of Transfer Learning along with the user-chosen architecture of Deep Learning Convolutional Neural Network (CNN). In the following, we refer to the proposed framework, developed for the purposed of crack detection as KrakN. 

\begin{figure*}[ht!]
    \begin{center}
        \includegraphics[width=0.75\textwidth]{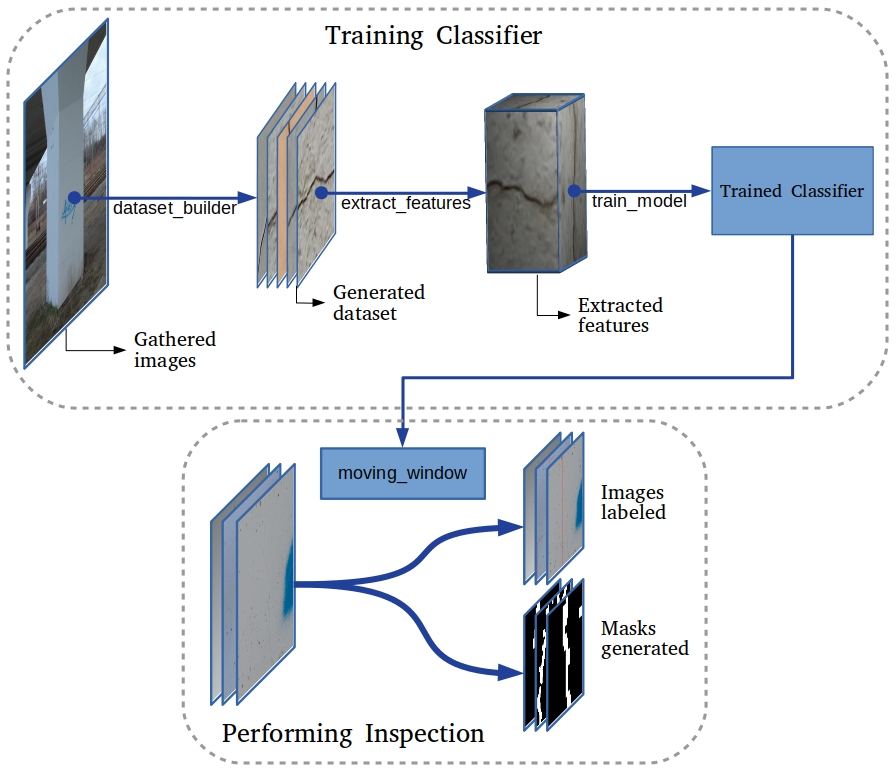}
    \end{center}
    \caption{Workflow in KrakN framework. Classifier training (upper panel) is required to introduce new type of defect into the database. Classification during inspection (lower panel) is based on trained model.}
    \label{fig:workflow}
\end{figure*}

The overall description of workflow in KrakN framework, including methods in their working order, is shown in Fig.~\ref{fig:workflow}. Note that the entire workflow is a two-step process. First, more labour intensive part -- classifier training will be done only once per infrastructure maintenance unit or when a new defect case will be added to the classifier. The classifiers obtained in this way can then be used throughout all of the departments without the need of the initial data. The second part, where the classifier is used, can then be repeated for each inspection carried out by the Bridge Inspector to support his work.

The comparison between previously described approach based on object detectors and the approach used in KrakN framework can be found in Tab.~\ref{tab:comparison}. 

\begin{table*}[ht!]
    \begin{center}
        \begin{tabular}{|>{\columncolor[gray]{0.8} \centering\arraybackslash}m{0.3\textwidth}|>{\centering\arraybackslash}m{0.3\textwidth}|>{\centering\arraybackslash}m{0.3\textwidth}|}
            \hline 
            &\cellcolor[gray]{0.8} \textbf{Object detectors} &  \cellcolor[gray]{0.8} \textbf{KrakN framework} \\ \hline
            \textbf{Image resolution} & Limited & As camera resolution \\ \hline 
            \textbf{Initial training data demand} & High data demand, label and object position on image required & Low demand with use of Transfer Learning, only labels required  \\ \hline 
            \textbf{Dataset building} & Manual & Automatic, included in the framework \\ \hline 
            \textbf{Training hyperparameters} & User-set & Managed by the framework \\ \hline 
            \textbf{Accuracy} & Limited by anchors number & Limited by camera resolution \\ \hline 
        \end{tabular}
    \end{center}
    \caption{Comparison of approach based on object detectors and the approach utilized in KrakN framework.}
    \label{tab:comparison}
\end{table*}

\subsection{Convolutional Neural Networks in object detection}

Convolutional Neural Networks have been developed mainly for the purpose of image classification. The basic feature of these algorithms, which allowed them to achieve exceptional results in image classification, is their spatial awareness \cite{Li2018,rosebrock2017deep}. It means that during the training, the Network learns not only the characteristic features of the given object but also their specific spatial location. To achieve that, they use locally connected convolutional layers with trainable weights. The image resulting from the convolution operation is also reduced in size allowing the Network to learn the features of the image starting from local features in the initial layers to global ones as the depth of the Network increases.

\begin{figure}
    \begin{center}
        \includegraphics[width=3in]{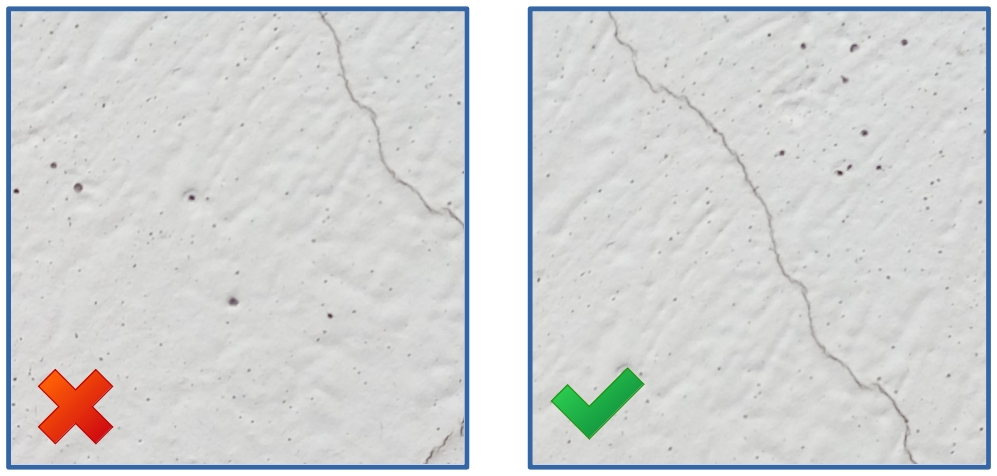}
    \end{center}
    \caption{Incorrect (left) and correct (right) placements of a defect for training dataset.}
    \label{fig:placement-training}
\end{figure}

The above characteristic has a major impact on the data that the algorithm is working on.  In particular, for building a dataset, the attention should be paid not only to the presence of a given object in the image but also to its spatial location. Due to this necessity, the position of the classified object should be consistent in all of the training samples as well as when using the algorithm during the classification. This principle is demonstrated in Fig.~\ref{fig:placement-training}, where both images contain the sought defect, but only one of them should be used for training the algorithm in order to achieve the best performance.

\begin{figure*}
    \begin{center}
        \includegraphics[width=0.75\textwidth]{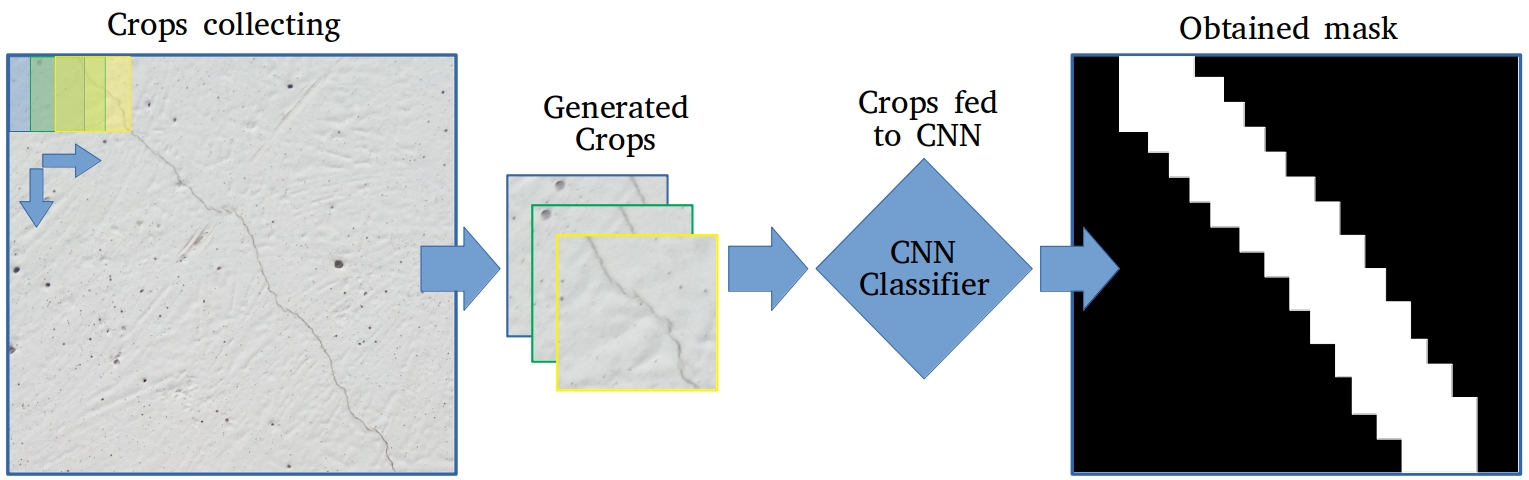}
    \end{center}
    \caption{Preview of the sliding window technique}
    \label{fig:sliding-window-preview}
\end{figure*}

The spatial awareness of Convolutional Neural Networks is very important when they are utilized as an object detector using the sliding window technique, as shown in Fig.~\ref{fig:sliding-window-preview}. To minimize the number of cases in which the image crops used for feeding the image classification algorithm omit the sought defect resulting in false negative classification, a certain overlap of the crops should be applied. While it boosts the accuracy of the prediction, it also significantly increases the number of images fed to the classification algorithm, which significantly extends the run time. Therefore, in order to obtain satisfying accuracy while limiting overall computation time, by the general rule of thumb, the overlap value should be kept between 0.50 and 0.75 of the input image size. This value is set to 0.60 by default in the software package presented in this work, but can also be modified manually by the user.

As can be seen in Fig.~\ref{fig:sliding-window-preview}, the result of the sliding window algorithm is a set of classified, overlapping crops that can later be used to form a mask for the initial image. Such masks can be utilized for trimming out defects from the input for further analysis and cataloguing or enriching IFC models with additional data \cite{Sacks2018}. In the software package presented in this work, both the generated masks and original images with marked defects can be obtained.

\subsection{Convolutional Neural Network architecture}\label{sec:cnn-architecture}

Each of the Convolutional Neural Networks is characterized by its architecture, defined as a specific arrangement of the network layers. In the early stages of development of these algorithms, CNNs had no more than ten trainable layers. It was caused mainly by the limited computing power of computers at the time. The examples of such architectures are LeNet \cite{Lecun1998} (three convolutional layers) and AlexNet \cite{Krizhevsky2012} (five convolutional layers). Despite the shallow structure of these networks, resulting in their limited accuracy, they are still used in a range of image classification implementations in which there is limited computing power, \emph{e.g.} single board computers for IoT applications.

\begin{figure*}
    \begin{center}
        \includegraphics[width=\textwidth]{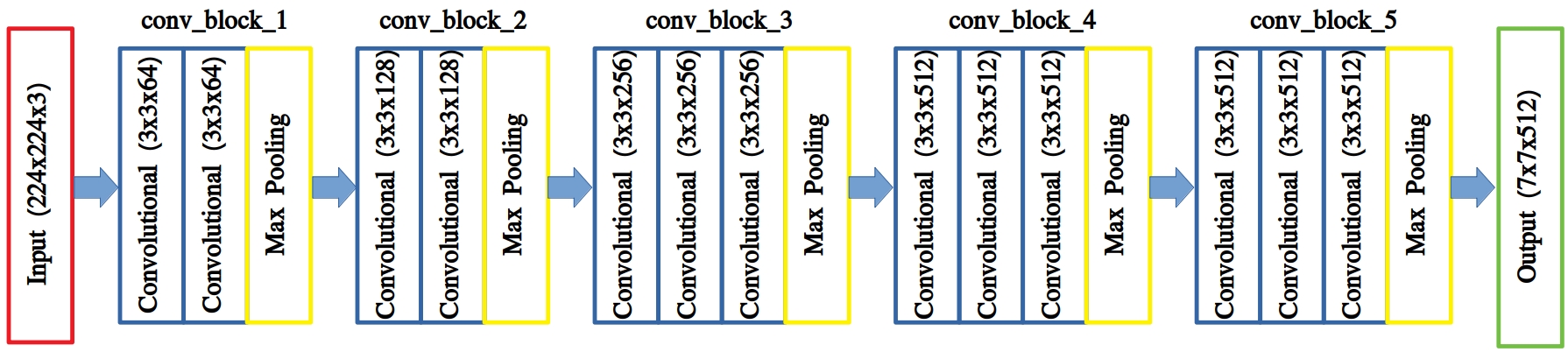}
    \end{center}
    \caption{VGG16 network architecture used in the presented study.}
    
    \label{fig:vgg16-architecture}
\end{figure*}

Modern CNN architectures are much deeper, meaning they consist of more trainable layers than their initial implementations. Current state-of-the-art Neural Networks architectures (\emph{e.g.} ResNet-152) reach over 100 layers and are capable of human-like accuracy in image classification. However, for the purpose of the article, a simpler yet still effective architecture of VGG16 network, consisting of 16 trainable layers was adopted. Its architecture with the shape of filters in each layer is shown in Fig.~\ref{fig:vgg16-architecture}. One should note that in the scope of the Transfer Learning methods used in the articles case study, the presented architecture does not contain the final layers of a network with fully connected layers and a classifier layer used for the dataset that the network was initially trained on.

It is also important to notice that, unlike object detectors like SSD and YOLO, conventional CNNs used for the sliding window approach accept only fixed image sizes as input, as required by the network architecture. It is especially apparent while building own training datasets, for the training data dimensions have to be directly linked to the chosen CNN model. In the KrakN software package, it is possible to set the dimensions of the dataset images manually, but by default, the dimensions are set to $224\times224\times3$, where the third number is the number of image channels.

As shown in Fig.~\ref{fig:vgg16-architecture}, the main building blocks of the network consist mainly of convolutional layers in which the feature extraction and training take place. In addition, the spatial dimensions of the input image after passing through the CNN are reduced from $224\times224$ to $7\times7$. The total number of parameters exceeds 14 million in the transfer learned part of the network and exceeds 130 million, considering the last, fully connected layer and a classification layer.

\subsection{Benefits of Transfer Learning}\label{sec:tl-principles}

\begin{figure*}
    
    \begin{center}
        \includegraphics[width=0.75\textwidth]{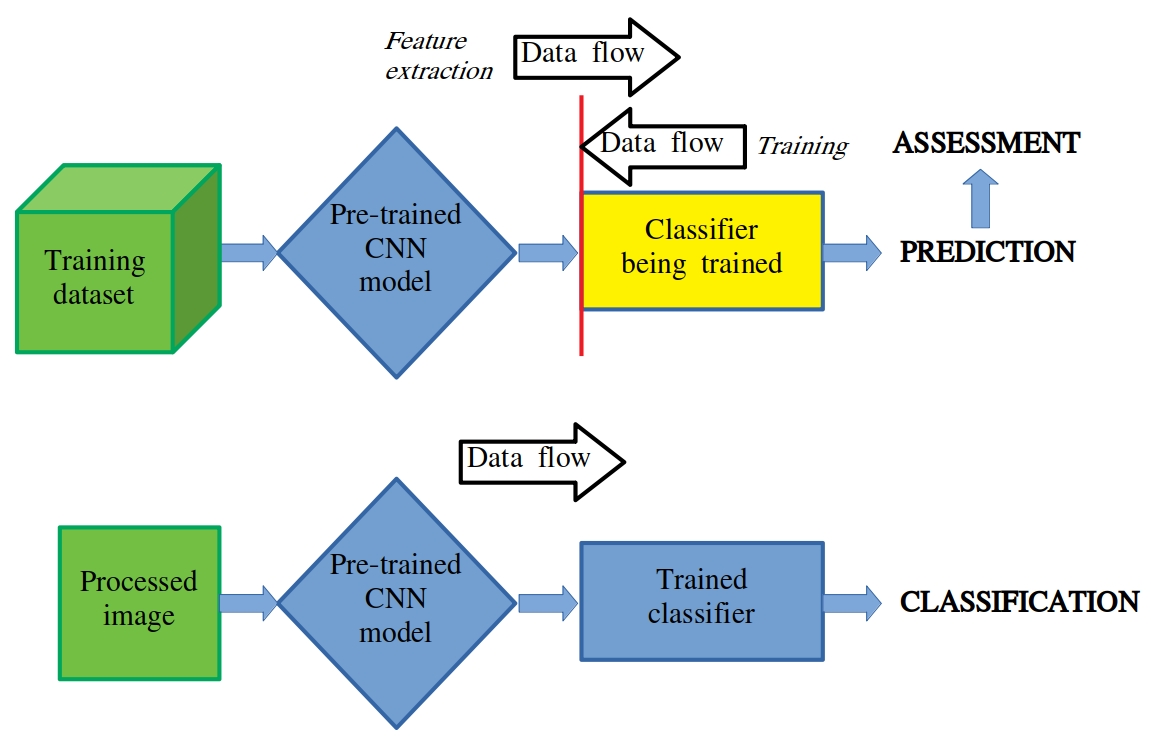}
    \end{center}
    \caption{Transfer-learned Network during training (top) and image classification (bottom).}
    
    \label{fig:transfer-learned-network}
\end{figure*}

Transfer Learning is a technique that allows the use of a network trained on one dataset as a feature extractor for another dataset, thus reducing the number of network layers that need to be trained again for the new set of images. It is based on the assumption that regardless of the specific objects sought in the image, their general characteristics will be similar across all datasets and the extracted features in the convolutional part of the network will be universal to some extent. The only layers that are subjected to training with this approach are the last fully connected and the classification layers that learn specific features of the new dataset. This process, along with the direction of data flow in the consecutive steps of training and image classification, is shown in Fig.~\ref{fig:transfer-learned-network}. Utilizing this approach it is also possible to reduce or extend the total number of classification classes for the CNN to learn, as compared to the number of output classification classes of the initial network. In the KrakN package, the number of classes was reduced from 1000 to 2.

There is a number of benefits of this approach that will potentially enable the use of Machine Learning methods for the local governmental units involved in infrastructure maintenance. The first one is the significant reduction of computing power demand, since in this approach only the last fully connected layers are subjected to training during backpropagation of the data through the network. This results in training time multiple times shorter and less resource heavy as compared to training whole CNN from scratch. 

Another benefit is that, according to studies \cite{Li2018}, using Transfer Learning requires much less training data. What's more, by using models developed for popular contests, \emph{e.g.} the ImageNet challenge \cite{ImageNet}, there is a certainty that they were initially trained on vast, correctly and carefully labelled data.

Finally, the proposed framework introduces the ability to quickly exchange the CNN core network if new model trained on a database of closer resemblance to the current application is available. Moreover, the framework can be used with an entirely new, deeper algorithm that utilizes more complex architecture for better accuracy results. The only caveat is that, in the latter case, the user has to be aware of the possible change in the input data dimensions or algorithm performance.

\section{Dataset collection and framework implementation}\label{sec:implementation}

This section provides a description of the process of data collection, dataset building using computer aided methods and network training with Transfer Learning in the form of a case study. Code flowcharts demonstrating crucial elements of the presented software package are also presented along with examples of the algorithm's output for the ease of the implementation in the practical applications. The presented framework is then tested on the dataset of thin cracks. Dataset utilized in this case study is available in the KrakN repository~\cite{KrakN}.

\subsection{Programming environment and libraries}

KrakN software package~\cite{KrakN} presented in this paper was developed using popular open-source programming environment. Table~\ref{tab:software-elements} presents the main programming environment and libraries divided into sections with their version number used in the project necessary to reproduce the obtained results . All of the software, libraries and methods utilized in the project are either open source or developed by the authors for the purpose of the project. It is also possible to utilize KrakN package using online cloud environments such as Google Colaboratory to minimize computing power requirements. The appropriate version of the software package in the form of executable notebooks is included in the project repository. The repository also includes detailed instructions for installing required dependencies. One should note that the algorithms described in the article were developed and tested on Ubuntu 18.04 LTS GNU/Linux system and the libraries may vary between the operating systems. One should also note that services such as Google Colaboratory, eliminate the burden of configuring local programming environment altogether.

\begin{table*}
    \begin{center}
        \begin{tabular}{|>{\centering\arraybackslash}m{0.155\textwidth}|>{\centering\arraybackslash}m{0.185\textwidth}|>{\centering\arraybackslash}m{0.187\textwidth}|>{\centering\arraybackslash}m{0.18\textwidth}|>{\centering\arraybackslash}m{0.155\textwidth}|}
            \hline 
            \rowcolor[gray]{0.8} Programming environment & Machine Learning & Math, matrix management & Utilities, database management & System management \\ 
            \hline
            Python 3.6 & Tensorflow (1.12)
            & Numpy (1.16) & H5py (2.9) & Os (built-in) \\ 
            & Keras (2.2)
            & Opencv (4.0) & Pickle (built-in) & Imutils (0.5) \\ 
            & Scikit-learn (0.22) & Random (built-in) & Progressbar (2.5) & PyGame (1.9) \\ 
            \hline 
        \end{tabular}
    \end{center}
    \caption{Elements of the programming environment used for developing and testing KrakN software package. Detailed description of the installation procedure can be found on the project web page~\cite{KrakN}.} 
    \label{tab:software-elements}
\end{table*}

\subsection{Collection of training, testing and validation data}

There are many free datasets available on the internet that can be used to train the algorithm, including Civil Engineering dataset SDNET2018 \cite{Dorafshan2018a} for crack detection. However, they usually do not focus on a single type, or in the case of cracks -- single width range of damage -- what can often can lead to mislabelling of particular defect type in the scope of local regulations. For this reason, for the sake of completeness of dataset preparation, a full process of image gathering and labelling is presented.

As a case study for dataset building, a heavily cracked bridge pillar was used. Moreover, a single test subject for training set was purposely selected in order to verify the framework ability to generalize knowledge, whereas for evaluation set, images from multiple cases and cameras were collected. Another premise for using single structure for training set is to simulate the worst case scenario, where the bridge inspector has very limited, homogeneous dataset for training. Pictures for the dataset were taken with a Sony Alpha DSLR-A500 digital camera and 18-55 mm f/3.5-5.6 focal length lens in good lighting conditions at a distance of 20-30 cm from the surface of the pillar. All of the pictures were taken by hand. In result, over 900 pictures with a resolution of $4248\times2850$ px, covering most of the pillar surface were obtained. In practical applications, photos for dataset or inspection itself can be obtained using UAV vehicle \cite{Zhong2018} or self-driving rovers equipped with CCTV cameras as in \cite{Cheng2018}, for further limiting the labour intensity while increasing the possibility of reaching hard to inspect places during the infrastructure inspections. It should be stressed out that for the task of classifying the images taken during the inspection, a different camera can be used than when collecting the dataset. This makes it possible to use widely available cheap cameras present in smartphones (see  Fig.~\ref{fig:moving-window-detection}) for further lowering the deployment cost.

The obtained photos contained not only the sought defects of thin cracks, but also concrete pores, dirt, background surface and other features that could be sought during the actual inspection, but for the sake of the case study, only cracks and background surface were taken into consideration in dataset building as two classes for labelling.

\begin{figure}
    \begin{center}
        \includegraphics[width=4in]{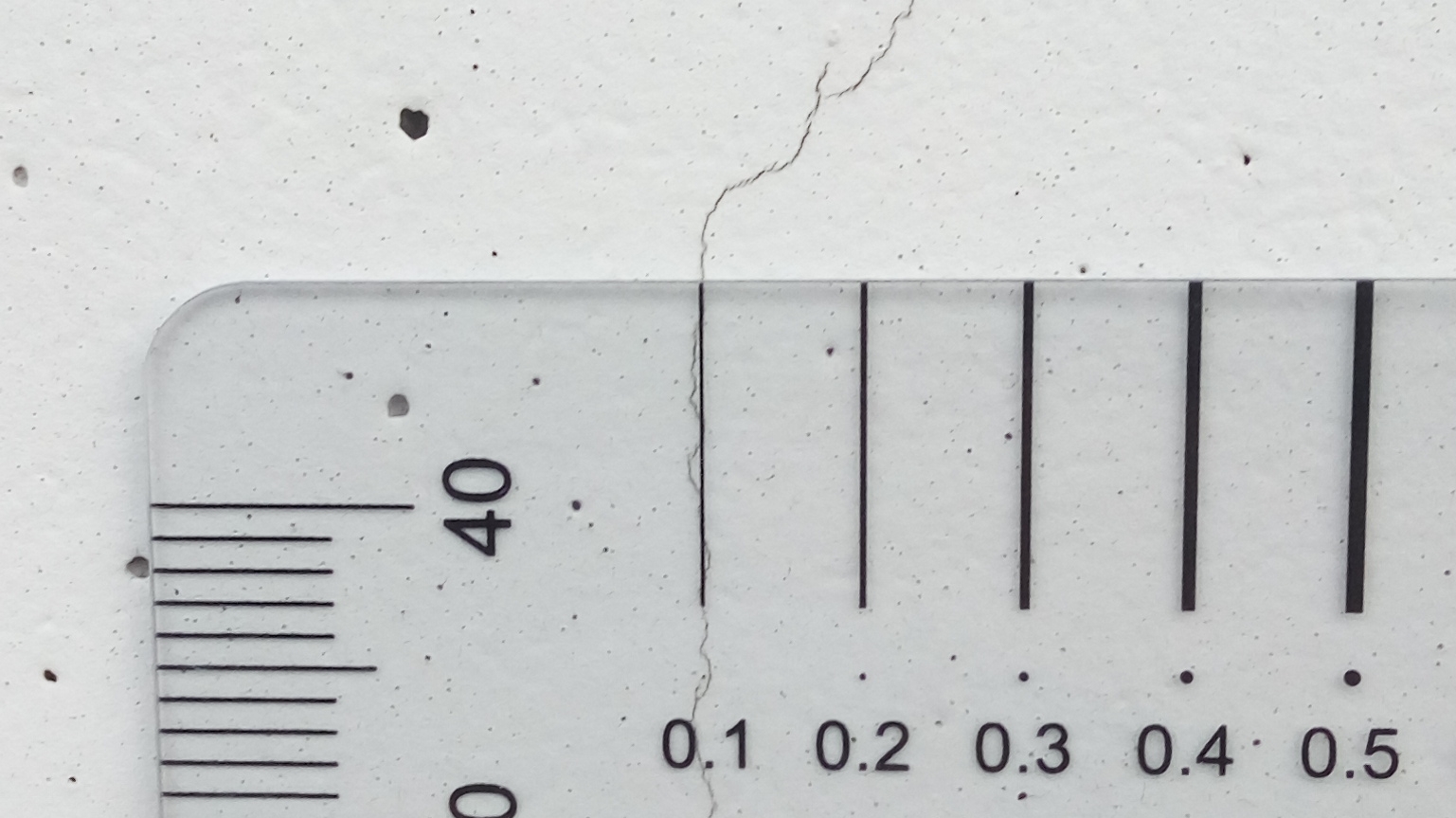}
    \end{center}
    \caption{An example measurement of crack captured for the dataset}
    \label{fig:crac-example}
\end{figure}

To measure the method performance only cracks with the width in range below 0.2 mm were considered. Such cracks are often omitted during the inspections. Example of a crack used in the presented case study can be seen in Fig.~\ref{fig:crac-example}.

\subsection{Script-aided dataset building}\label{sec:dataset-building}

After gathering a set of required images for the dataset, they must be processed into the labelled crops of a fixed size, also referred to as the data points. This process is both time and labour consuming, as the dataset building is often described as the hardest and most costly step of object detector training. There is also no existing out-of-the-box software which would significantly reduce the labour intensity of this process, since the data points for each of the CNN architecture may vary in demanded size.

\begin{figure}[ht!]
    \begin{center}
        \includegraphics[width=2in]{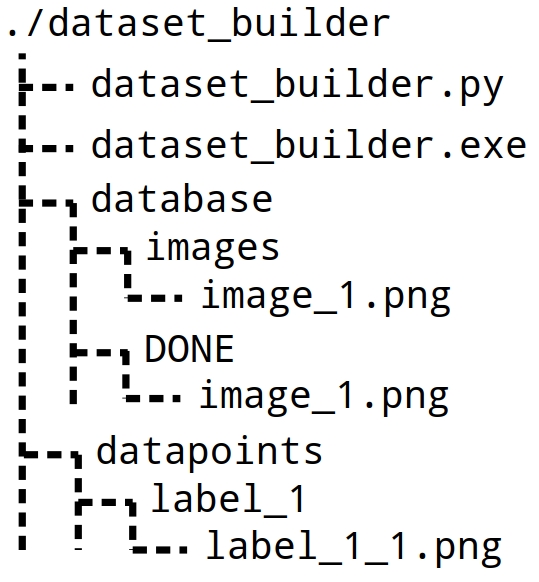}
    \end{center}
    \caption{The \code{dataset_builder} directory structure. Subdirectory \code{database} contain input images and subdirectory \code{datapoints} contains crops gathered in subdirectories corresponding to the labels.}
    \label{fig:builder-directory}
\end{figure}

For this reason, KrakN software package includes computer program created to improve the division of a dataset into labelled data points, called \code{dataset_builder.py}, also provided in compiled \code{dataset_builder.exe} format for Windows users. In order to use it, a certain directory structure presented in Fig.~\ref{fig:builder-directory} should be maintained. Note that due to Google Colaboratory limitations, this is the only part of the workflow that cannot be done with this service.

The main directory, containing file \code{dataset_builder.py}, is further divided into two subdirectories. The \code{database} directory contains unseen images in \code{images} folder and already processed images in \code{DONE} folder. The \code{datapoints} directory is where the user should place self-named folders with names corresponding to desired class labels names. These folders will store data points in the form of image crops obtained from the input photos. If the folders are not provided, one dataset class will be created by default.

\begin{figure*}
    \begin{center}
        \includegraphics[width=\textwidth]{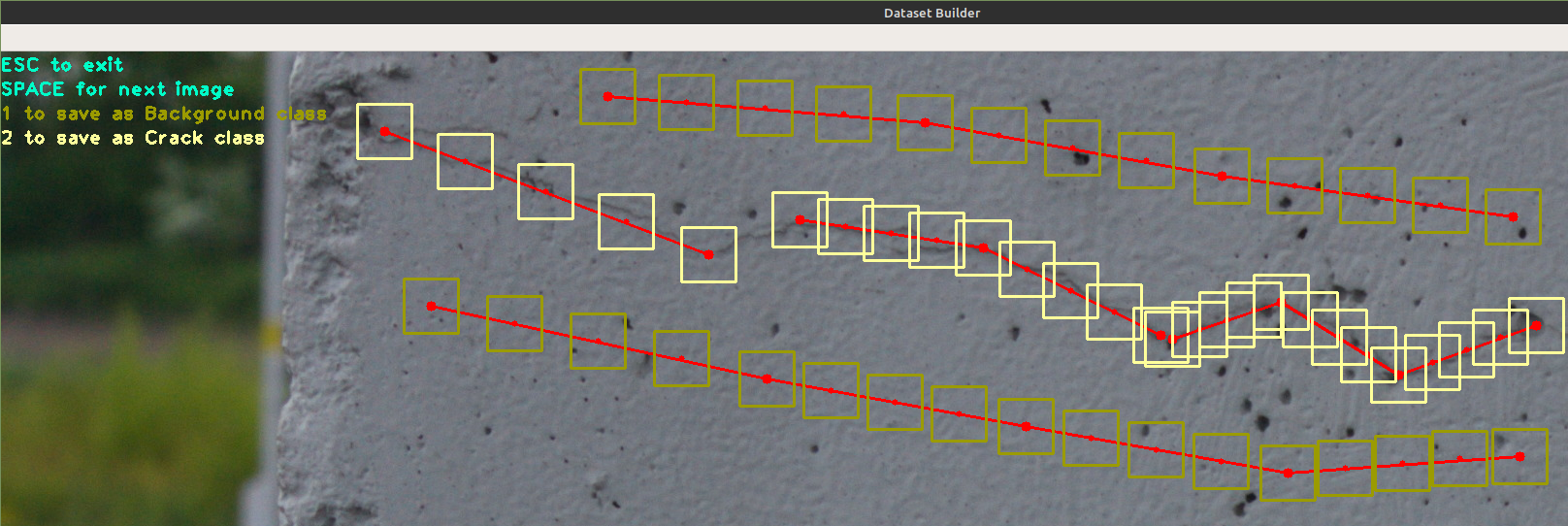}
    \end{center}
    \caption{Dataset\_builder interface with marked datapoints for both considered classes}
    \label{fig:builder-interface}
\end{figure*}

To extract the data points from the input image, the mouse is used to select the path on the image, along which the frames will be extracted. The user then selects one of the previously set class name that is loaded automatically into the program. By default, each of the path section is used to extract 5 crops with default dimensions of $224\times 224$ px. Those values can be altered by user by editing \code{dataset_builder.py} file. Since the algorithm uses the sliding window approach, it does not automatically compensate for different scale of the sought defect. Therefore, the user has to enter the desired scale factor in the first step of the workflow, considering the defect size in the input image. The value will be saved in the output crop file names to be used later in the algorithm operation. In the presented case study the scale factor was set to 2, due to the small size of the considered defect. An example of the  dataset building script interface and its operation is shown in Fig.~\ref{fig:builder-interface}.

As seen in Fig.~\ref{fig:builder-interface}, the \code{dataset\_builder.py} script allows for quick extraction of datapoints from the input image, as over 60 correctly labelled crops were extracted and placed in separate folders from the single input image. For the purpose of the case study, basing on over 900 input images, over 8 thousand datapoints for each class were extracted from the dataset using \code{dataset_builder.py}. The whole process of datapoint extraction took less than 4 hours to complete.

\subsection{Transfer Learning implementation details}

As discussed in Section~\ref{sec:tl-principles}, training classifier using Transfer Learning consists of two phases --- the feature extraction and the actual training of the last classifier layer (see Fig.~\ref{fig:transfer-learned-network}). While both of these elements can be done in one continuous cycle, it is often more beneficial if the feature extraction and the classifier training are done separately. This way it is possible to train the classifier quickly multiple times with different parameters or regression models without the need to extract the features from the dataset each time, as the features will be saved into a separate file. It is also worth noting that for big datasets the feature extraction part will usually be a more time-consuming process than the classifier training itself.

\begin{figure*}
    \begin{center}
        \includegraphics[width=3.2in]{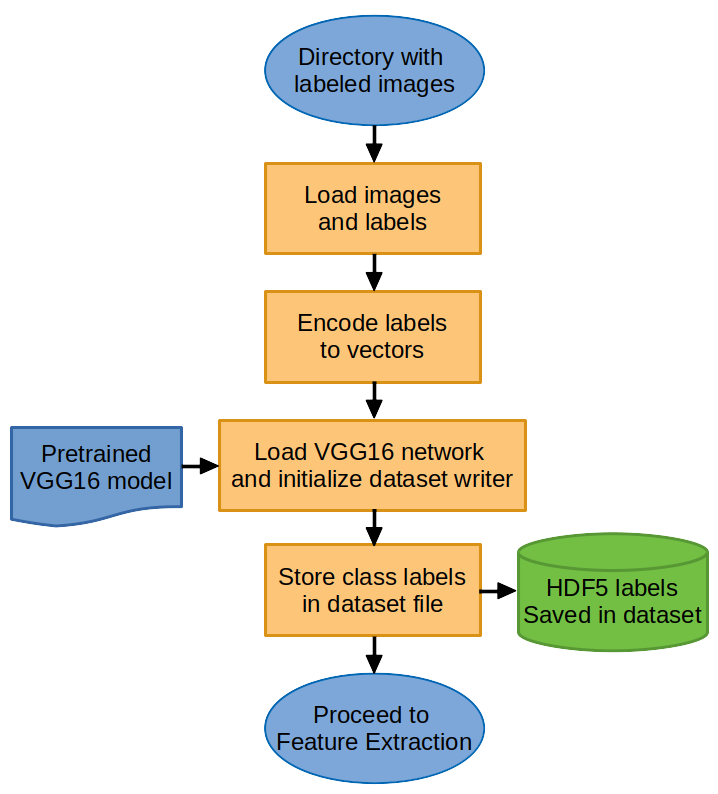}
    \end{center}
    \caption{Flowchart of labels preprocessing in \code{extract_features.py} script}
    \label{fig:flowchart1}
\end{figure*}

\begin{figure*}
    \begin{center}
        \includegraphics[width=3.2in]{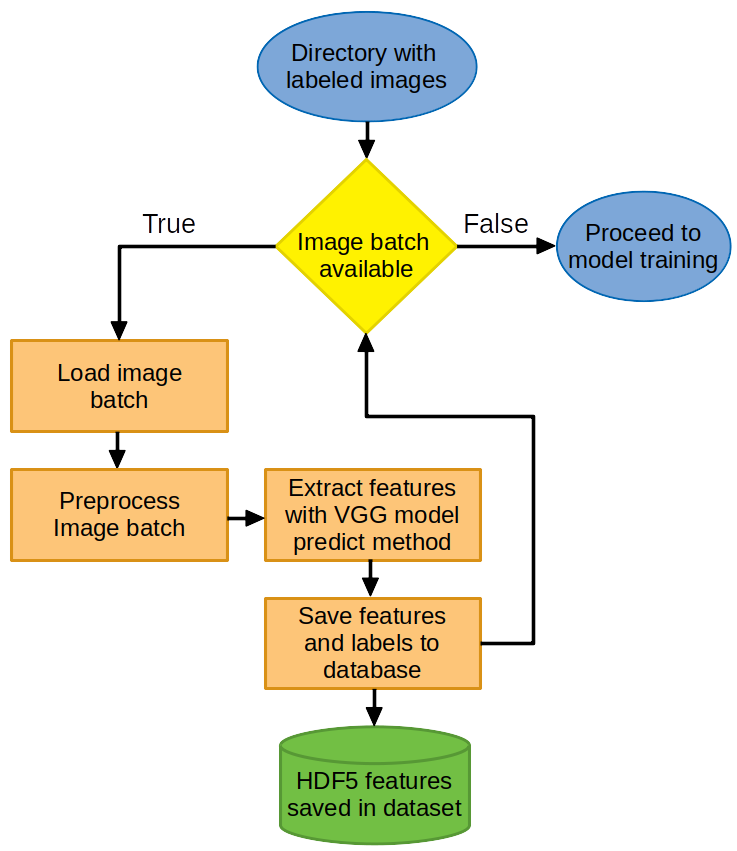}
    \end{center}
    \caption{Flowchart of feature extraction in \code{extract_features.py} script}
    \label{fig:flowchart2}
\end{figure*}

In KrakN software package, the feature extraction process is implemented by \code{extract_features.py} script. The script can be run on a local machine, but for better performance due to GPU enabled computations in the cloud, the use of Google Colaboratory service is suggested. The key elements of the workflow can be found in Figures~\ref{fig:flowchart1} and \ref{fig:flowchart2}.

In the first phase of the feature extraction presented in Figure~\ref{fig:flowchart1}, class labels derived from folder names are loaded. Next, the labels are encoded to vectors using \code{sklearn.preprocessing} library. Then script loads desired CNN architecture without its last layer as \code{include_top} parameter in script is set to \code{False} by default. One should note that in the presented case, the initialized weights come from network trained on \code{ImageNet} dataset. The last parts initialize an object that will store the extracted features as a dataset for classifier training in \code{HDF5} file. Note that in the script, a parameter that indicates the dimensions of the extracted features coincides with the dimensions of the output of the network as seen in Fig.~\ref{fig:vgg16-architecture}. After initializing label encoder and dataset writer, the features are extracted and saved to a single file in a series of batches in the following step.

The core part of the feature extraction if presented in Figure~\ref{fig:flowchart2}. The main loop of the algorithm loads the image batch in the dimensions expected by the CNN and then preprocesses it according to CNN specific input. In the last step, the preprocessed images are accumulated in a single batch and the feature extraction takes place with the use of the previously loaded VGG16. Finally the features with corresponding labels are saved in the output \code{HDF5} dataset.

Execution of the script results in the creation of the \code{HDF5} file in \code{database} folder that contains all of the features and labels of the input dataset. Its size strongly depends on the number of input images. Because, unlike input images, the data is not compressed in any way, its size will usually be greater than the size of the images combined. In the presented case study, the size of the features extracted from over 16 thousand images exceeds 3 GB. With the use of Google Colaboratory service, the feature extraction of the case study dataset with \code{batchSize} parameter set to 128 took under 3 minutes, while on the local machine without utilizing powerful GPU unit, it would take up to 6 hours.

\begin{figure}
    \begin{center}
        \includegraphics[width=3.2in]{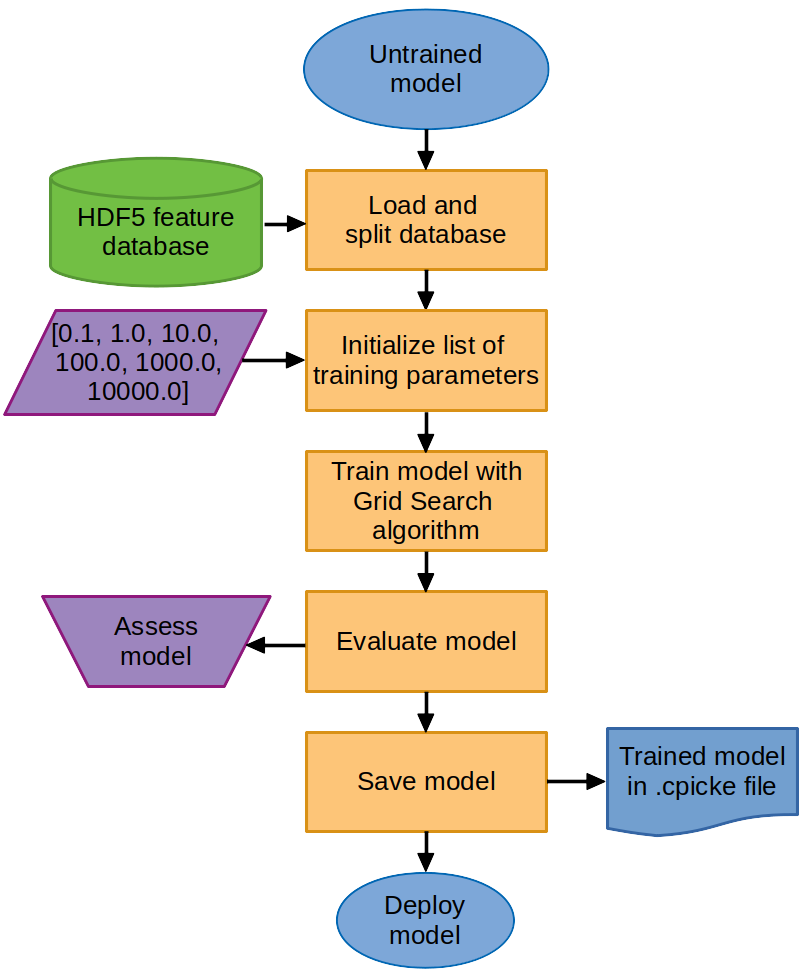}
    \end{center}
    \caption{Training model for deployment with \code{train_model.py} script}
    \label{fig:flowchart3}
\end{figure}

As already discussed in Section~\ref{sec:tl-principles}, Transfer Learning enables us to utilize the pre-trained model and train only the last layer of the CNN. In the presented software, this process is implemented in \code{train_model.py} script and its core part is presented in Figure~\ref{fig:flowchart3}. In the described approach, we utilize a grid search algorithm over logistic regression models trained with various hyperparameters. This way, out of all of the trained models, the model showing the best performance can be automatically selected.

In the beginning, algorithm loads the previously extracted features and compute the number of data points for the training and evaluating the model. It is most common to use the values that split the database in a ratio close to 0.75 in favour of training data. Next, a list of training parameters is created. These values also are most commonly initiated on a logarithmic scale. After that, both the grid search algorithm and the logistic regression model are created and trained. Next, the models are evaluated on the remaining part of the dataset. The user can then assess the results of the training with shown metrics. In the last step, \code{KrakN_model.cpickle} file is created, and the best performing estimator is saved in its location. The training process for the case study using Google Colaboratory service, utilizing CPU computations only, took less than 20 minutes with the evaluation precision of 97\% and 98\% for Background and Crack classes, respectively.

\subsection{Sliding window algorithm implementation}

\begin{figure}
    \begin{center}
        \includegraphics[width=5in]{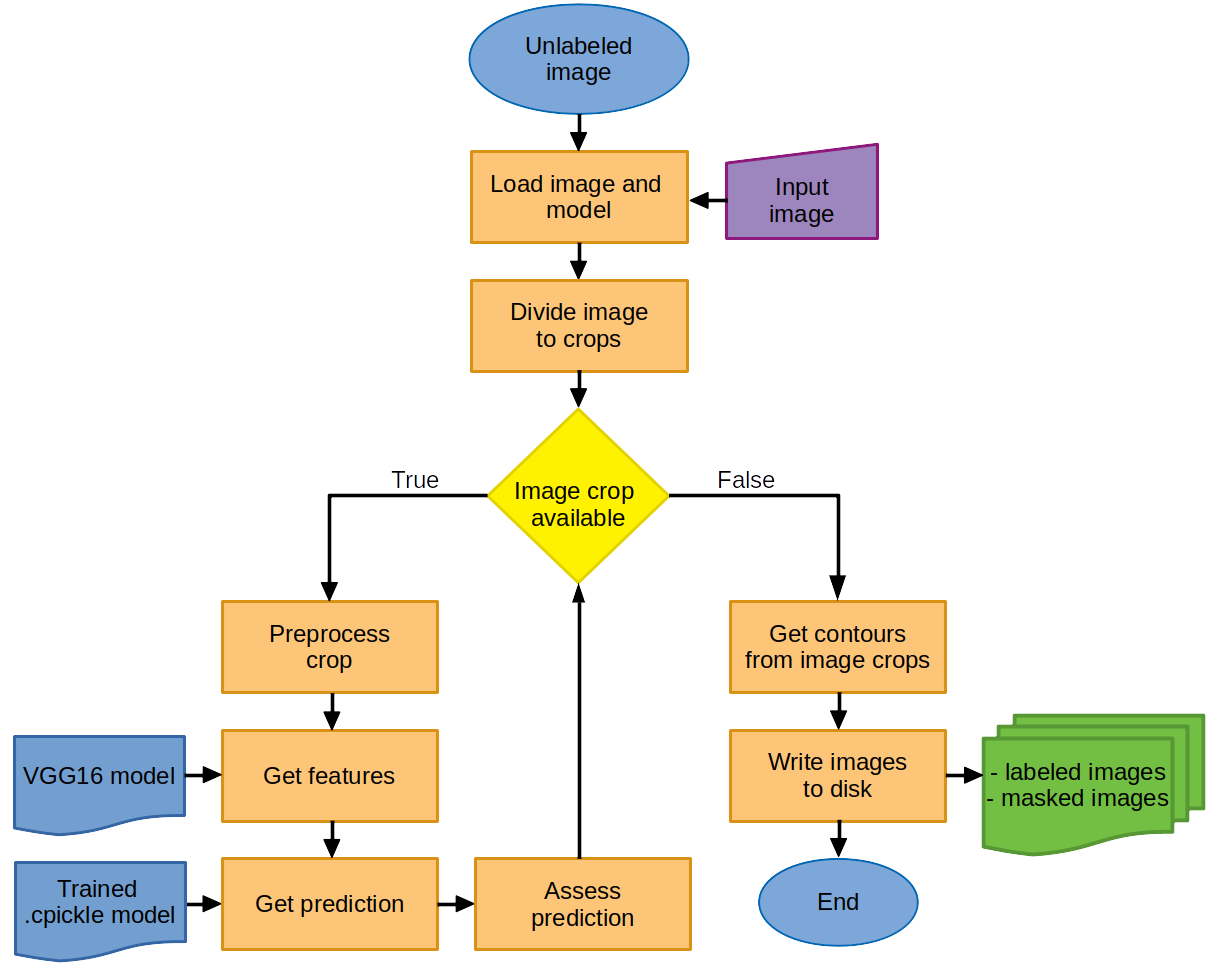}
    \end{center}
    \caption{Implementation of the sliding window in \code{moving_window.py} script}
    \label{fig:flowchart4}
\end{figure}

In order to use the previously trained classifier as an object detector, a sliding window algorithm has to be implemented. For the purpose of KrakN framework, such an algorithm was implemented in the \code{moving_window.py} script. The main loop of the algorithm is presented in Figure~\ref{fig:flowchart4}.

In the main loop, the image is cropped according to the scale of dataset, set while using \code{dataset_builder.py} and resized accordingly. As discussed in Section~\ref{sec:cnn-architecture}, the step by which the crops are selected is set to 0.60 of the crop size. Next, the input crop is preprocessed in the same way as it was for the purpose of feature extraction. Then, the previously trained classifier is used to predict the class of the selected crop. After prediction, the highest value is selected and compared to the confidence threshold in the prediction assessment step. This threshold is set to reduce the amount of false positive predictions. In the presented study, with only two classes, only a probability of over 50\% would result in assigning a crop to a class. This parameter is set by default to 0.95 but it can be changed by the user, based on the number of classes distinguished by the classifier.

After all predictions, the mask image files for every class are saved, as presented in Fig.~\ref{fig:sliding-window-preview} in Section~\ref{sec:methdology}. In the last step, based on the mask images, the output images with classes marked by red rectangles are generated for each class. Note that the rectangle markings are generated as bounding boxes for the mask images, so the classes in the output image can overlap --- this will be especially apparent for the background class. With the use of Google Colaboratory the whole process for $4248\times 2850$ px input image with over 2500 predictions per image, takes less than one minute. The example results of the algorithm are shown in Fig.~\ref{fig:moving-window-detection}.

\subsection{Workflow example results}

\begin{figure*}[ht!]
\begin{center}
    a) 
    \subfigure{\includegraphics[width=2in,height=2in]{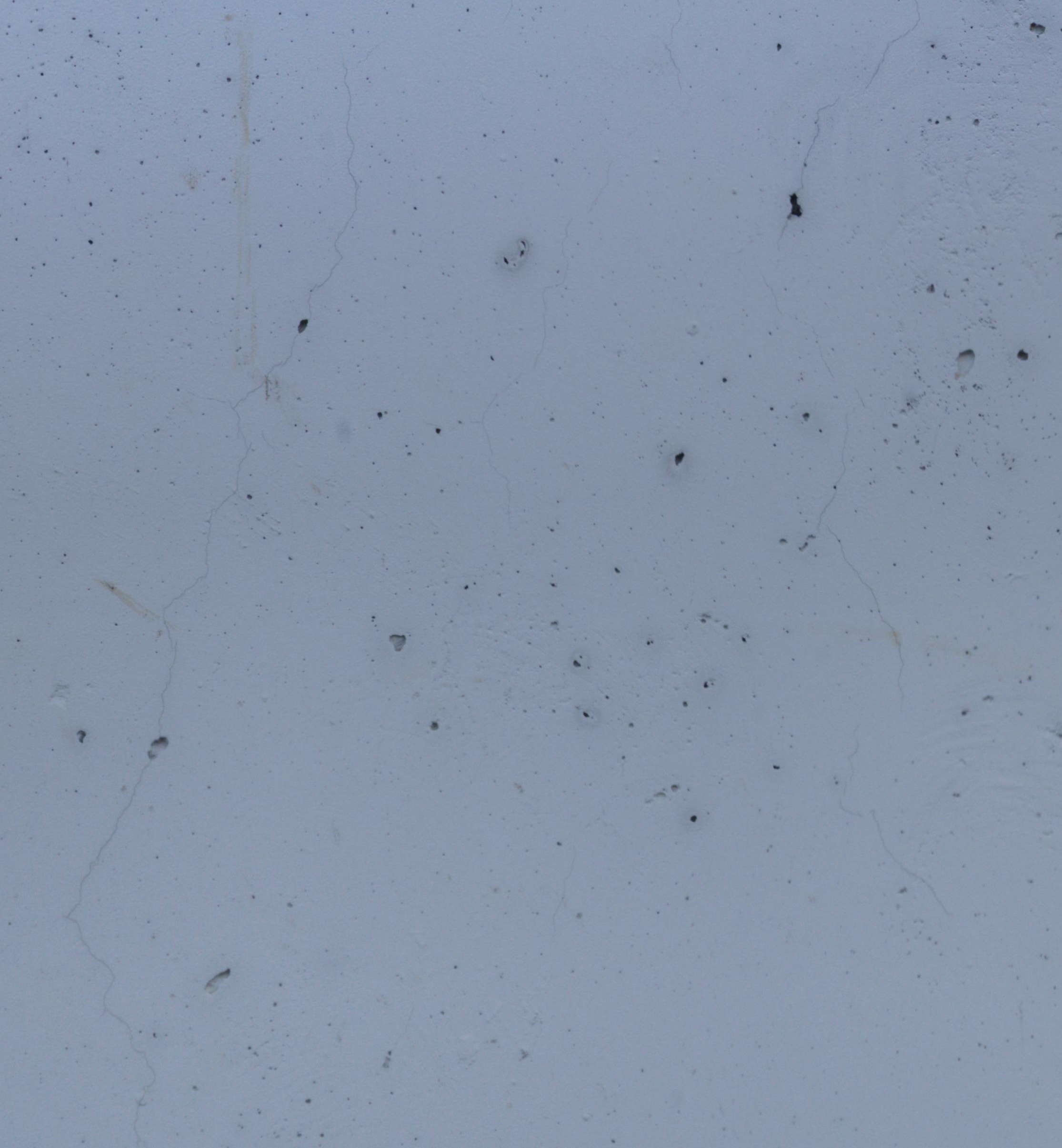}}\hspace{6pt}
    \subfigure{\includegraphics[width=2in,height=2in]{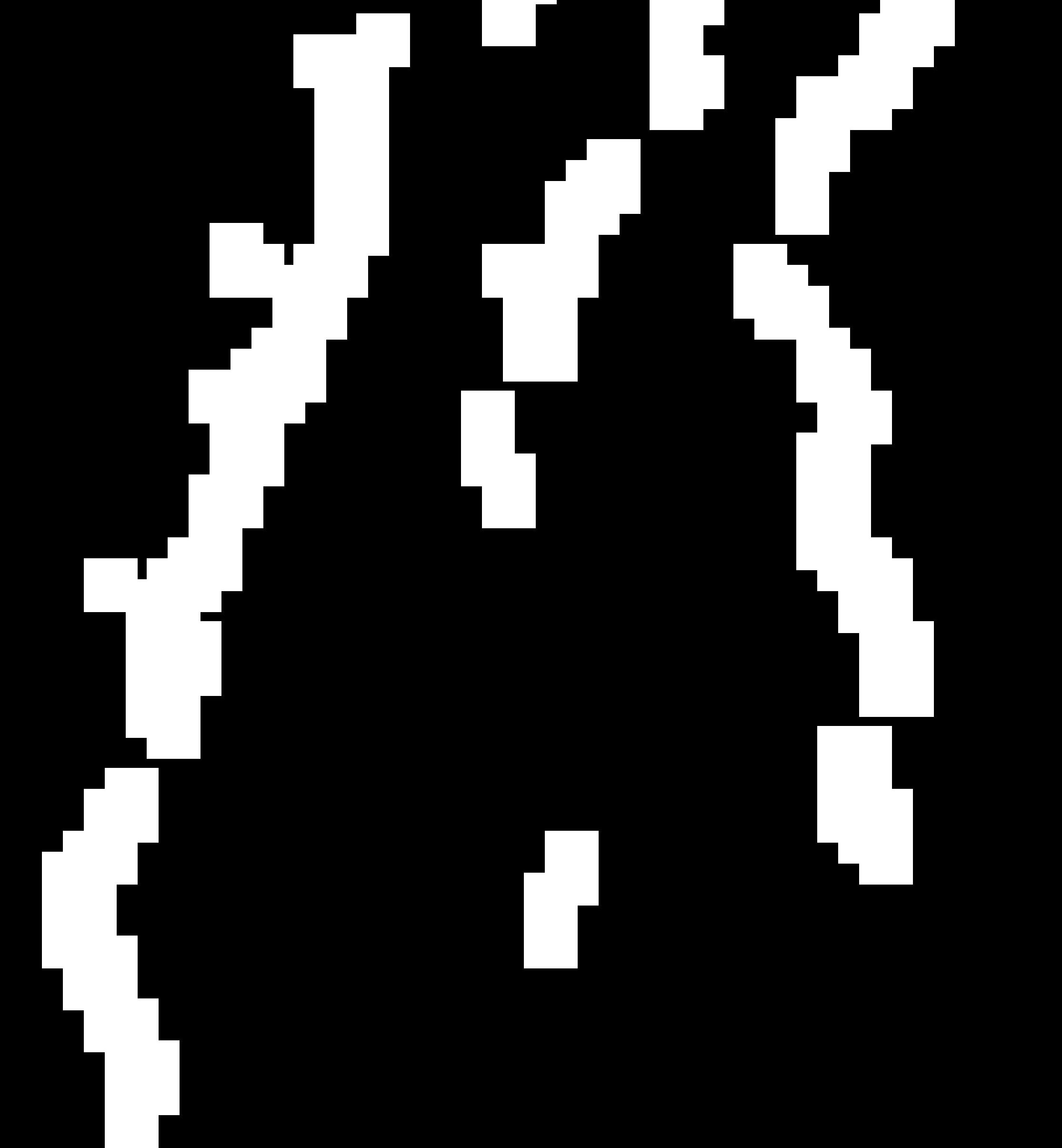}}\hspace{6pt}
    \subfigure{\includegraphics[width=2in,height=2in]{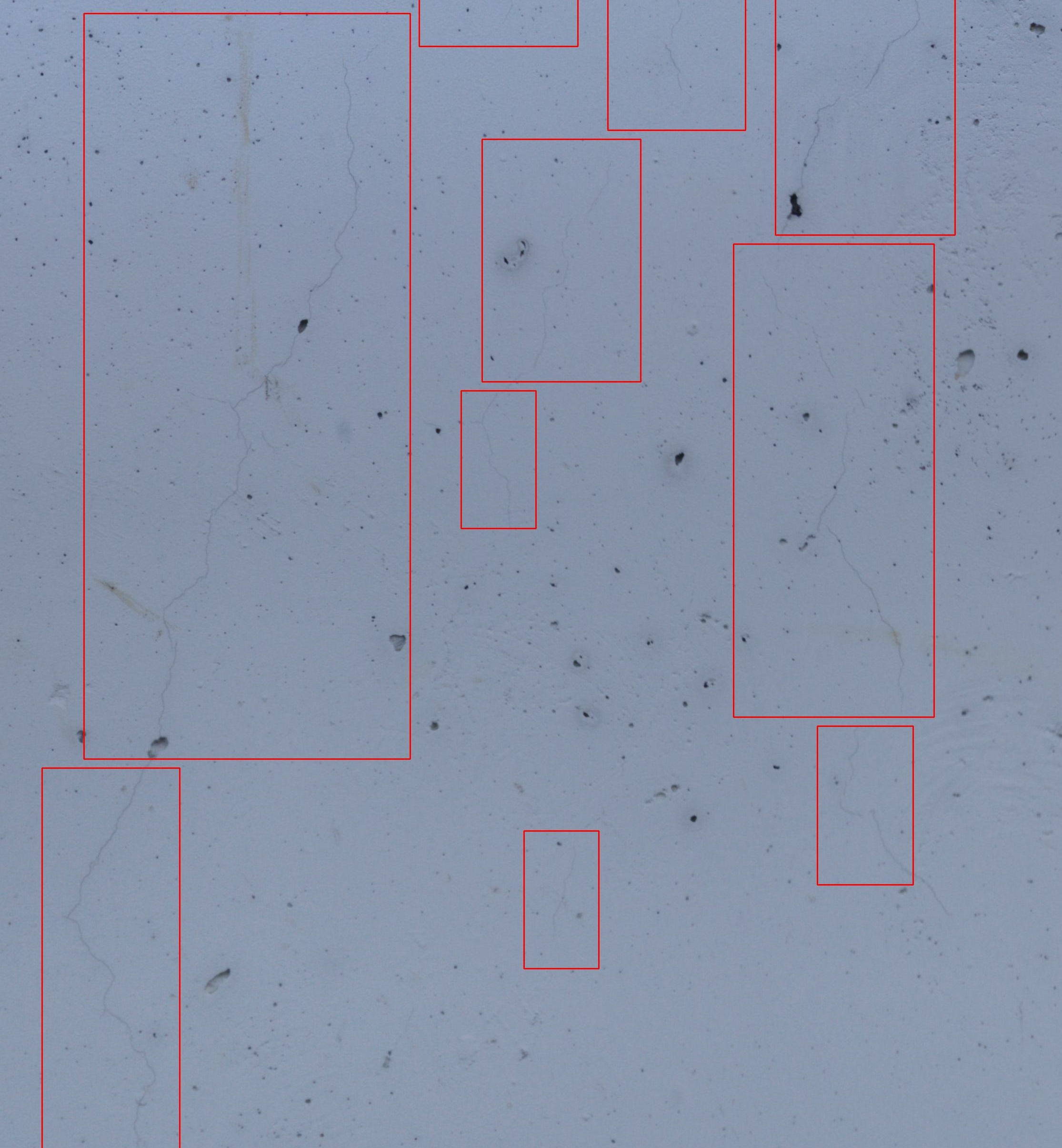}}\\
    b) \subfigure{\includegraphics[width=2in,height=2in]{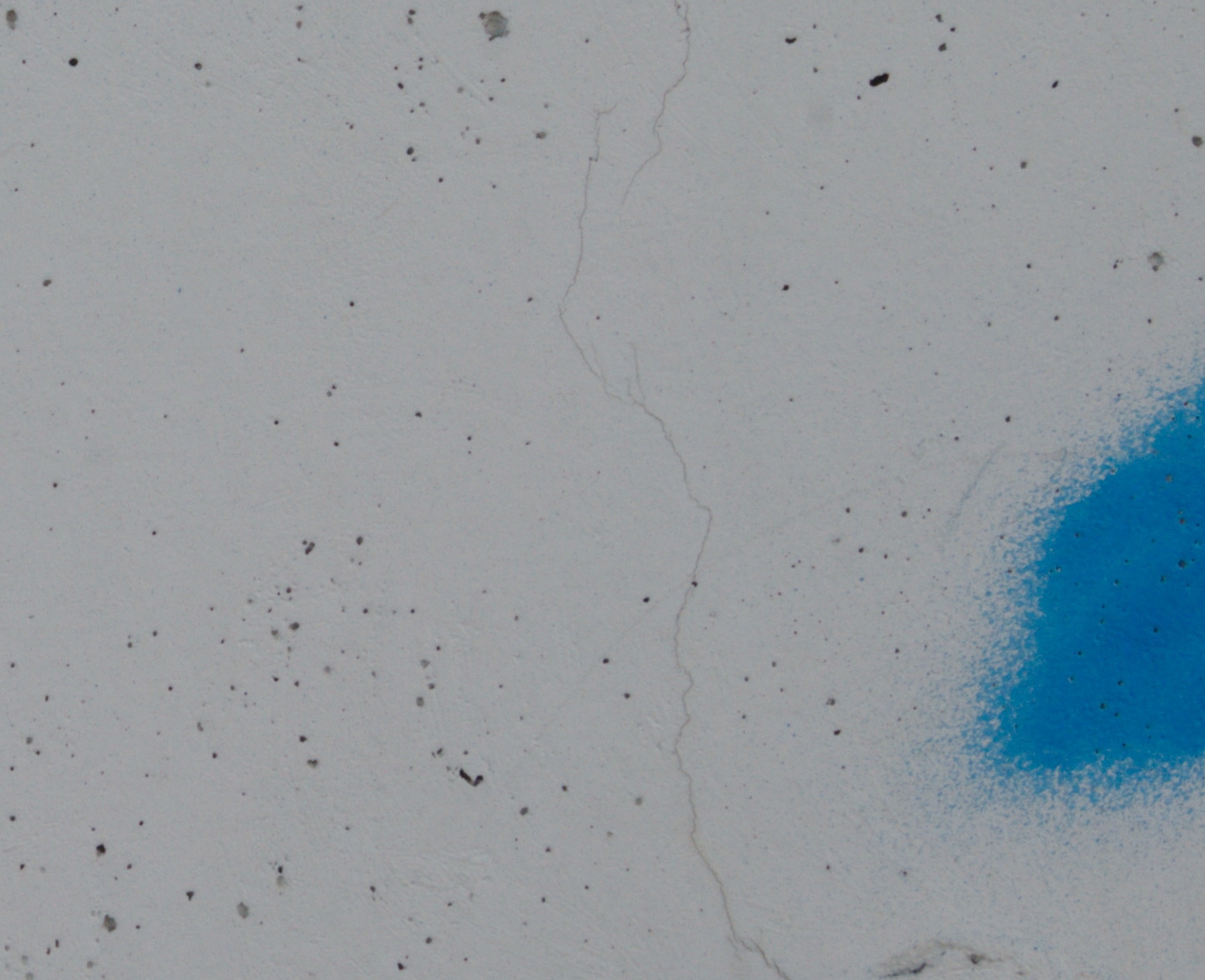}}\hspace{6pt}
    \subfigure{\includegraphics[width=2in,height=2in]{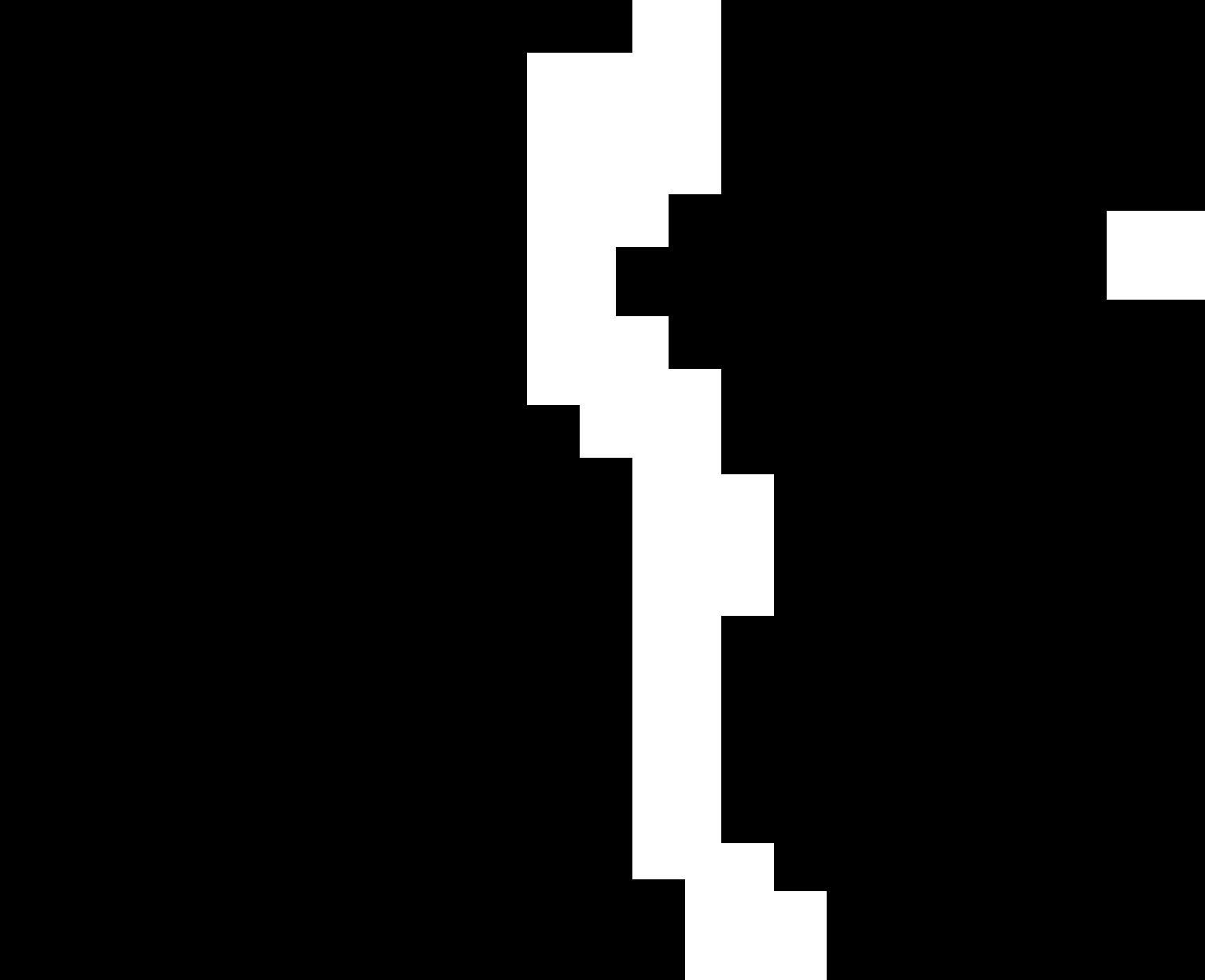}}\hspace{6pt}
    \subfigure{\includegraphics[width=2in,height=2in]{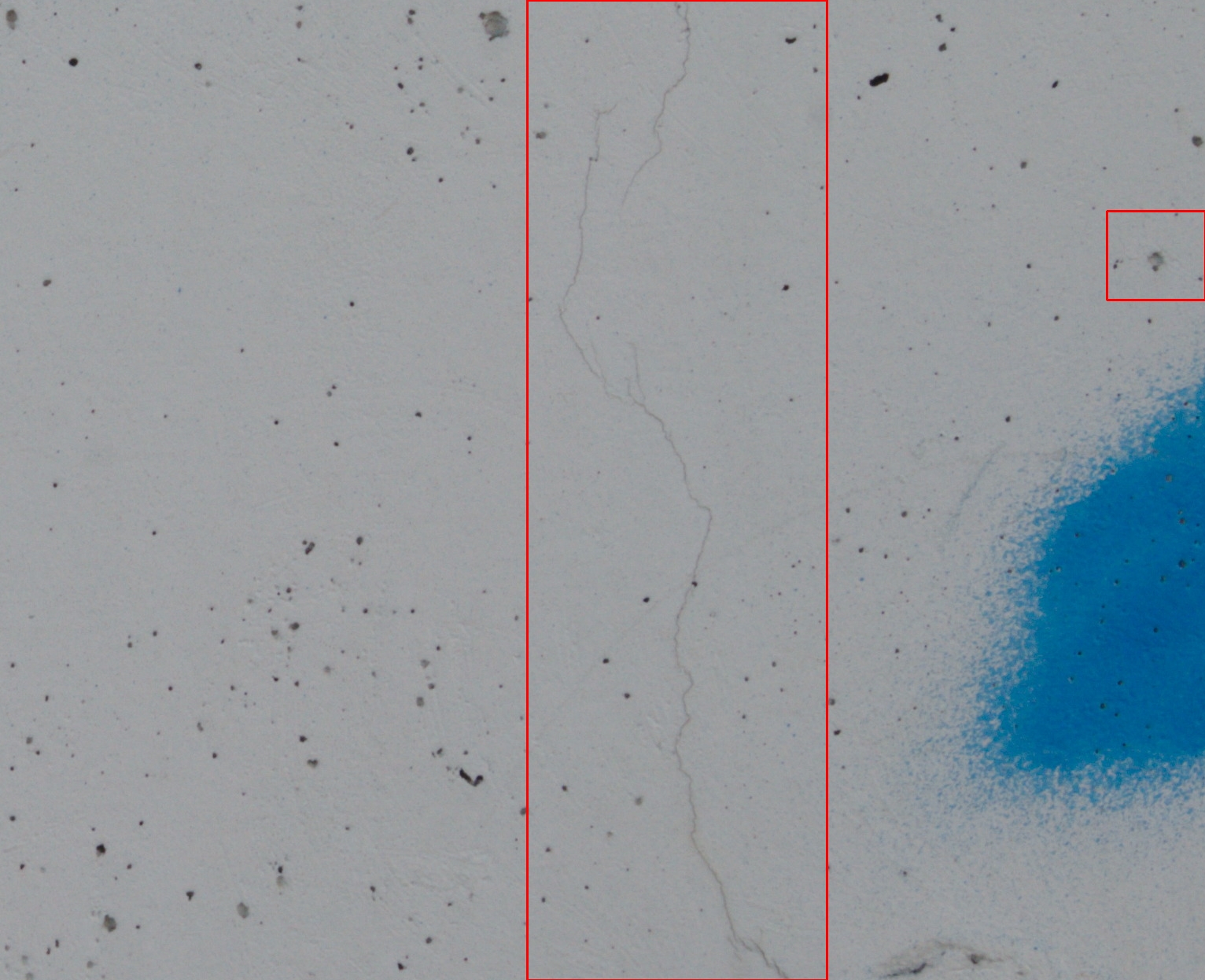}}\\
    c) \subfigure{\includegraphics[width=2in,height=2in]{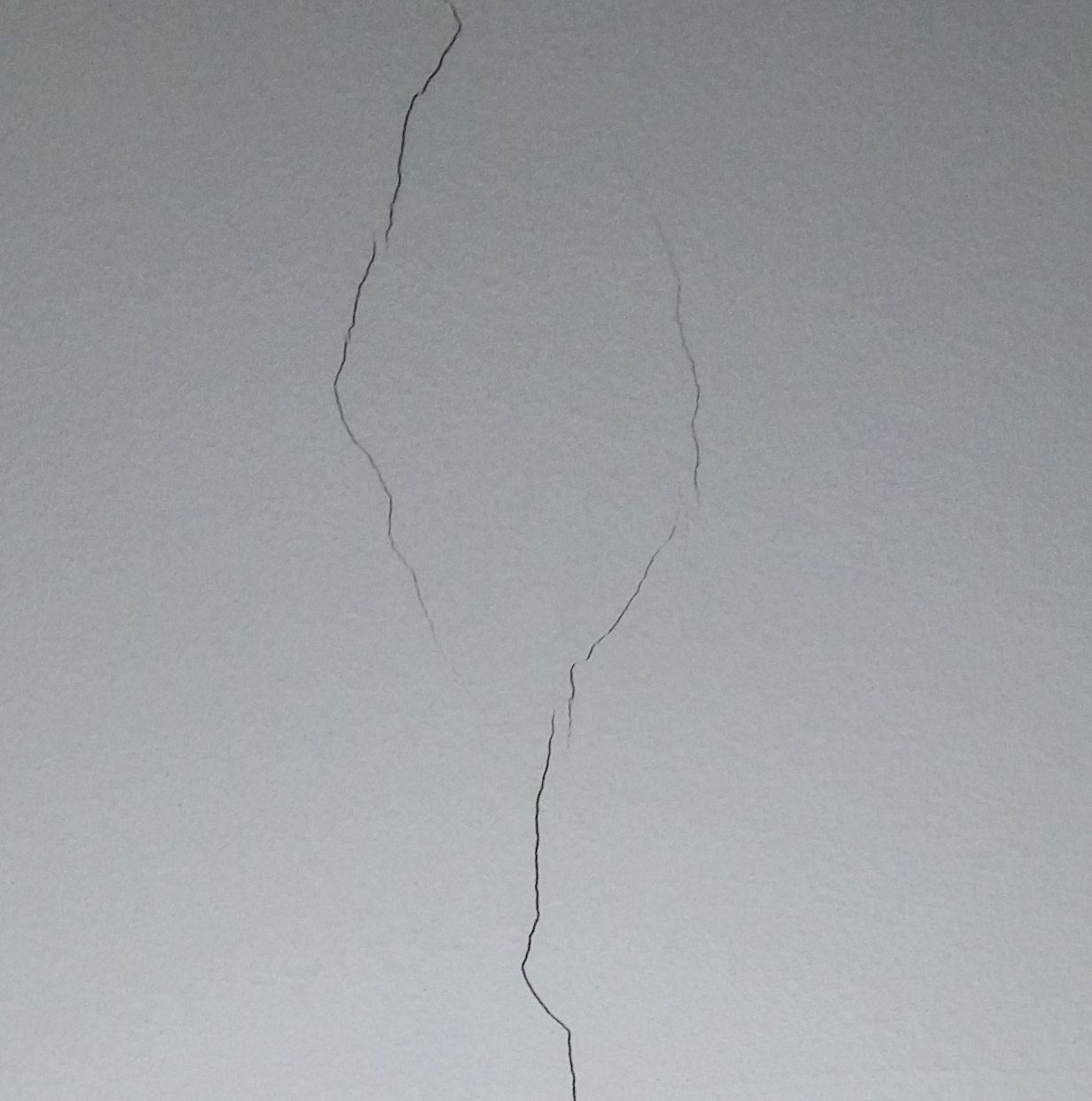}}\hspace{6pt}
    \subfigure{\includegraphics[width=2in,height=2in]{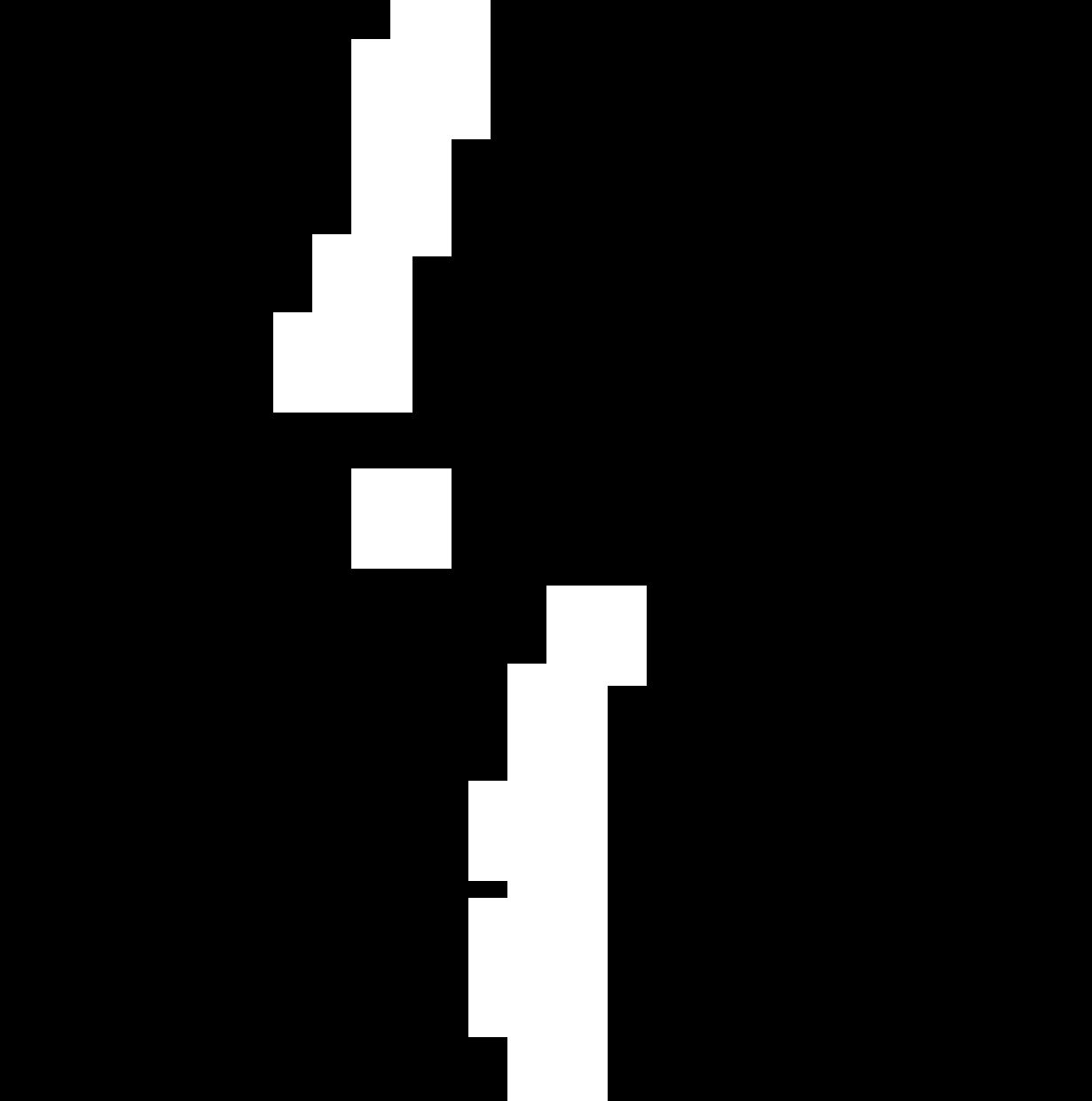}}\hspace{6pt}
    \subfigure{\includegraphics[width=2in,height=2in]{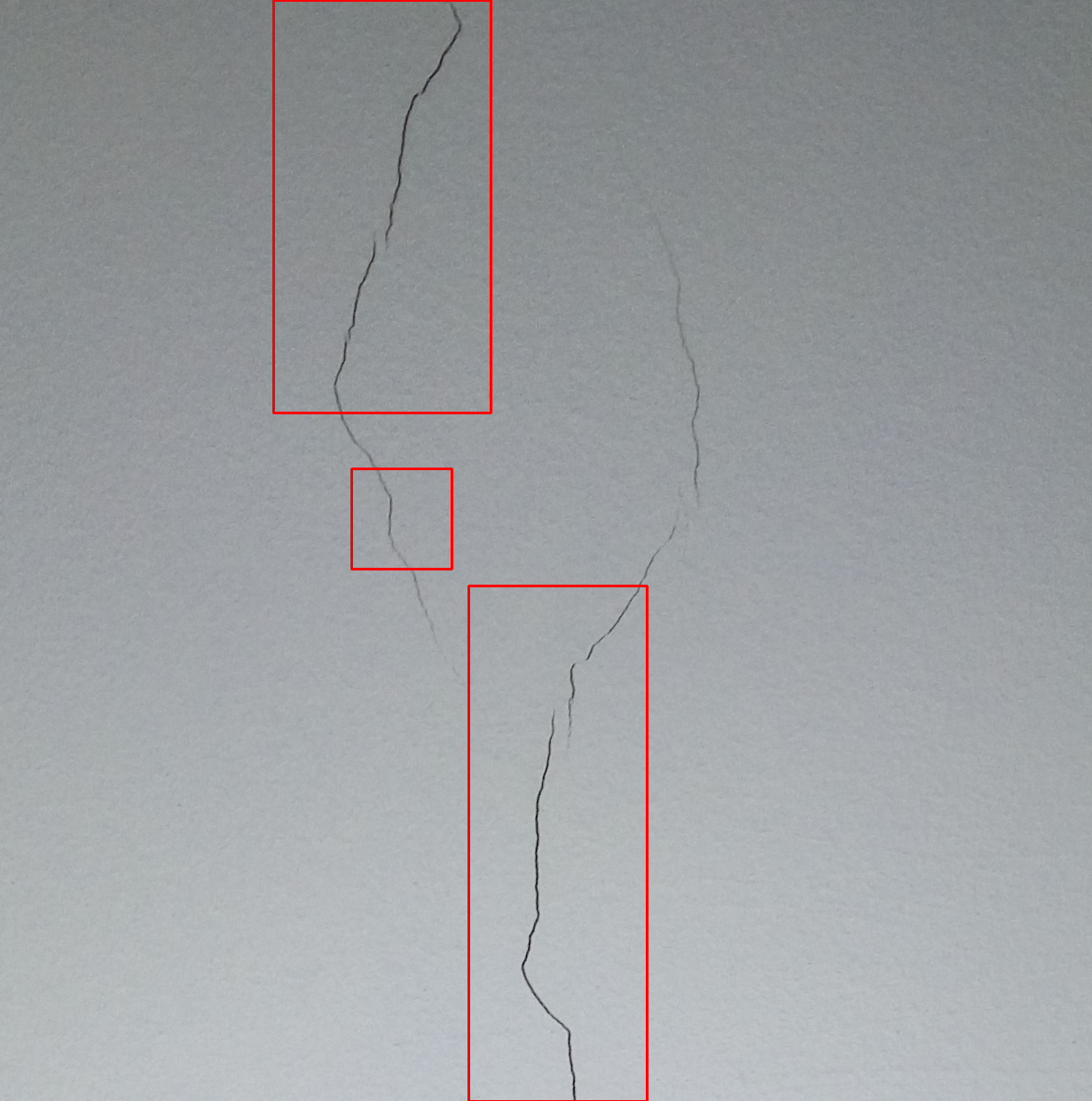}}
    
\end{center}

\caption{Examples of object detection outputs for \code{moving_window.py} script. a) Partial results of the script executed on the 120-megapixel orthomosaic generated on the test case bridge pillar, showing the ability of KrakN framework to detect defects hard to see with a naked eye on the large scale images. b) Close-up of the example results. Note that in the top right corner of the crack output image, there is a false positive prediction. c) Another example of the algorithm operation, albeit in this case a cheap (under 120 USD) smartphone camera was used. Despite the mediocre quality of the camera used and the observed surface significantly different from the training dataset, the algorithm maintained its ability to detect defects. However, the false negative predictions on the output image can also be seen.}
\label{fig:moving-window-detection}
\end{figure*}

Fig.~\ref{fig:moving-window-detection} provides an example of the results of the procedure implemented in \code{moving_window.py} script. As a result of the script, the output images are generated in the form of two-classes masks and images with the detected defects marked with rectangle bounding boxes. The generated masks can be later used for further defect management and cataloguing, while images with bounding boxes markings can serve as immediate indicators for the Bridge Inspector.

The individual cases depicted in Fig.~\ref{fig:moving-window-detection} show the possible results of the procedure. In the actual application of KrakN, the false positive to false negative ratio can be altered, according to the users' preferences, with the use of the \code{confidence_threshold} value, in order to minimize the threat of false negative or false positive predictions. In addition to the examples depicted in Fig.~\ref{fig:moving-window-detection}, more of the examples are included in the \code{examples} directory of KrakN repository.

\section{Framework benchmark}\label{sec:framework-benchmark}

To assess the reliability of the developed procedures, an experimental study, focused on the detection of thin cracks was conducted. Its main assumptions were to test the classifier on unseen data acquired on a variety of surface materials and coatings captured by various cameras and apertures under different lighting conditions and calculate the accuracy and recall of classifier predictions. These two evaluation metrics are used for obtaining the knowledge of the performance of the classifier. While accuracy gives the information about the overall performance of the algorithm, for problems with a high cost of false negative predictions such as infrastructure defects detection, recall metric is more important. It gives information about correct defect detection rate and can provide hints for increasing or lowering the confidence threshold value.

For the purpose of the experimental study, the classifier was fed with over 3 thousand unseen data points covering various types of surfaces comprising the case study classes of thin Cracks and Background. To enable the results of the experimental study to be reproduced, the evaluation set as well as the case study classifier were included in the project repository. The accuracy and recall were calculated using the following formulas:
\begin{eqnarray*}
    \text{Accuracy} &= & \frac{\text{true positives + true negatives}} {\text{total samples}},\\
    \text{Recall} &= & \frac{\text{true positives}}{\text{true positives + false negatives}}.
\end{eqnarray*}

Table~\ref{fig:cm-test} presents the results of the field test evaluation in a form of a confusion matrix. This data lead to the following evaluation metrics:
\begin{eqnarray*}
    \text{Accuracy} & = & 0.93,\\
    \text{Recall} &= & 0.93.
\end{eqnarray*}

\begin{table}
\begin{center}
\begin{tabular}{|>{\centering\arraybackslash}m{1.75in}|>{\centering\arraybackslash}m{1.75in}|>{\centering\arraybackslash}m{1.75in}|}
\hline 
 & \cellcolor[gray]{0.8} \textbf{Predicted Background} & \cellcolor[gray]{0.8} \textbf{Predicted Crack}\\ 
\hline 
\cellcolor[gray]{0.8} \textbf{Actual Background}  & 1577 (TN) & 125 (FP) \\ 
\hline 
\cellcolor[gray]{0.8} \textbf{Actual Crack} & 122 (FN)  &  1630 (TP) \\ 
\hline 
 & \cellcolor[gray]{0.8} \textbf{Total samples}  &  3452 \\ 
\hline 
\end{tabular}
\end{center}
\caption{Confusion matrix for the field test results where: TN (True Negative), FN (False Negative), FP (False Positive), TP (True Positive).}
\label{fig:cm-test}
\end{table}

It should be stressed out that the training data for the network was collected on a single structure, not related to evaluation set. Nevertheless, the developed procedures gained the ability to generalize knowledge across the different surfaces coatings, camera apertures, and lighting conditions while keeping high accuracy and recall of predictions. Note that in order to assess the algorithm properly, the confidence threshold was set to 0.50, therefore in practical applications where the confidence threshold is set according to the users preferences, the false positives to negatives category ratio can be different than shown in  Table~\ref{fig:cm-test}.

Despite the accuracy and recall results being lower than during the training of the algorithm, the results obtained are still much higher than those obtained during manual inspections (see \cite{Phares2004} and references therein). The most surprising results of the field tests were that the developed algorithm still scores over 50\% higher than human in detecting small surface defects despite being trained on a single infrastructure case and that the sought defects were nearly indistinguishable from the background by the naked eye. This demonstrates that the proposed method can be used in practical applications and provide robust method for supporting the automation of the process of infrastructure inspections.

\section{Comparative study}\label{sec:comparative-study}

For practical applications, aspects related to the ability of the method utilization in real-world scenarios are as important as the parameters describing their accuracy. For this reason, in this section we provide a comparative analysis, where both of those aspects of the KrakN software will be taken into account. The total time of the algorithm's training will be measured using machines with different parameters, as compared to similar algorithms trained from scratch. Then the performance of the algorithm on the evaluation set will be compared to current state-of-the-art methods. Lastly, a possibility of using KrakN as a multi-level infrastructure damage classifier (as in~\cite{Huthwohl2019}) will be presented.

\subsection{Comparison with network trained from scratch}

In order to perform comparative study highlighting the benefits of utilizing Transfer Learning in KrakN framework, a convolutional neural network of sequential architecture, similar to VGG16, was trained from scratch. The main assumptions of the experiment were to use the same dataset for training as in the case of Transfer Learning and then asses the network training time and accuracy as well as recall on the same evaluation set. The experiment was designed to check whether networks trained with Transfer Learning approach may have similar metrics and the ability to generalize knowledge from a narrow dataset to general case application as networks developed from scratch and to assess practical adaptability of networks trained from scratch in work-case scenarios.

\begin{table}[h!]
\begin{center}
\begin{tabular}{c c c}
     KrakN [CPU] & Conventional network [GPU] & Conventional network [CPU]  \\
     \hline
     \hline
     25 minutes & $\sim$10 hours & $\sim$444 hours (19 days)
\end{tabular}
\end{center}
\caption{Comparison of the training time for compared types of procedures.}
\label{fig:comparative-training-time}
\end{table}

\begin{figure*}
    \begin{center}
        \includegraphics[width=0.75\textwidth]{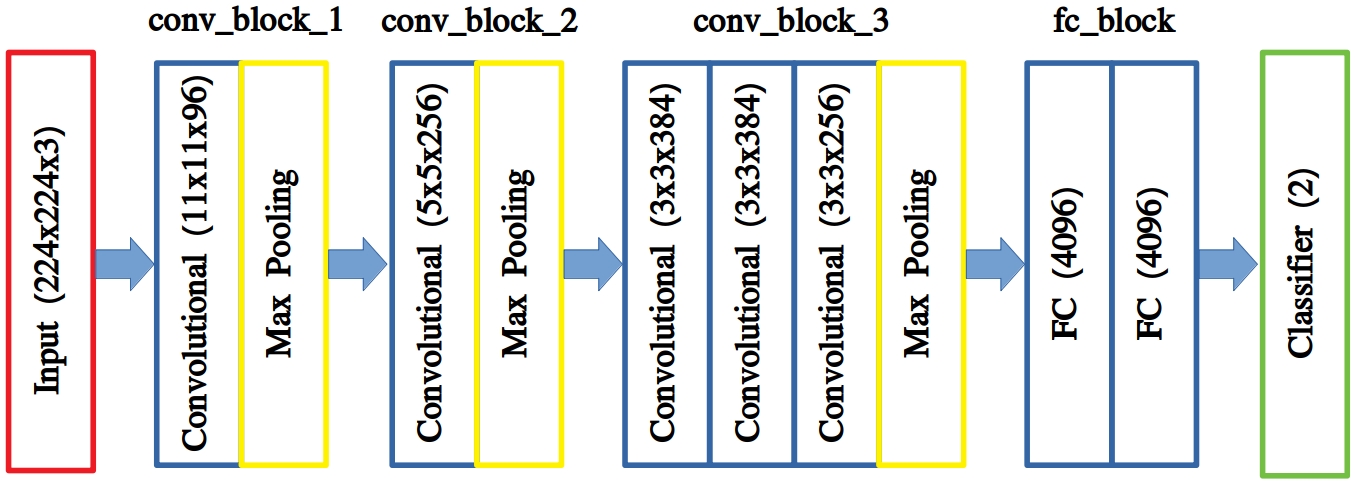}
    \end{center}
    \caption{Network architecture used for the comparative study.}
    \label{fig:comparative-network-architecture}
\end{figure*}

During the training of the network for comparative study, several architectures with the varying number of parameters and values of hyperparameters were tested. The average training time of one model using high-performance GPU (GeForce GTX 1080) was 10 hours, with the total training time for all tested models of over 50 hours under constant supervision. During the training, models were constantly evaluated for avoiding the CNN being biased towards one of the classes. The final network architecture used for the experimental study can be seen in Fig.~\ref{fig:comparative-network-architecture}.

As seen in Fig.~\ref{fig:comparative-network-architecture}, the architecture of network used for the comparative study is less extensive as compared to the CNN used for transfer learning (cf. Fig.~\ref{fig:vgg16-architecture}) During the training process, it was found out that model with less trainable layers was sufficient for the considered problem. Compared to the previously described model, it utilizes over two times less trainable parameters. This model was also included in the KrakN repository~\cite{KrakN}.

Considering the adaptability of the tested solutions in practical applications of infrastructure inspections performed by governmental units, it has to be noted that, as seen in Table~\ref{fig:comparative-training-time}, utilization of Transfer Learning significantly reduce computational resource requirements. KrakN framework trained with single-core Intel Xeon CPU provides 20-fold reduction in training time over network trained without any pre-training, even when high-performance GPU was used. The advantage of KrakN framework is more clear when only CPU units are available, as the conventional CNN training with the same CPU would take nearly 20 days of continuous computations. Furthermore, the values presented in Table~\ref{fig:comparative-training-time} show only one iteration of training for all of the methods. KrakN framework manages training parameters for the user, allowing for the algorithms training to be run only once, whereas conventional CNN generally require to be trained multiple times for parameters testing before the final deployment. Considering also that the tested CNN architecture had two times less parameters than VGG16, it makes KrakN approach an even more viable solution.

\begin{table}
\begin{center}
\begin{tabular}{|>{\centering\arraybackslash}m{1.75in}|>{\centering\arraybackslash}m{1.75in}|>{\centering\arraybackslash}m{1.75in}|}
\hline 
 & \cellcolor[gray]{0.8} \textbf{Predicted Background} & \cellcolor[gray]{0.8} \textbf{Predicted Crack}\\ 
\hline 
\cellcolor[gray]{0.8} \textbf{Actual Background}  & 1599 (TN) & 117 (FP) \\ 
\hline 
\cellcolor[gray]{0.8} \textbf{Actual Crack} & 103 (FN)  &  1630 (TP) \\ 
\hline 
\end{tabular}
\end{center}
\caption{Confusion matrix for the field test results}
\label{fig:cm-comparative}
\end{table}

Table~\ref{fig:cm-comparative} presents the confusion matrix obtained on evaluation set for network presented in Figure~\ref{fig:comparative-network-architecture} trained from scratch.
It lead to the following evaluation metrics:
\begin{eqnarray*}
    \text{Accuracy} & = & 0.94,\\
    \text{Recall} &= & 0.94.
\end{eqnarray*}

Comparing the results of training of both networks, it can be seen that the metrics of the network trained from scratch are only slightly better than in the case of transfer learning. Both networks have gained the ability to generalize knowledge across different datasets based on a limited amount of preliminary data collected from a single case in a comparable way. Both algorithms also obtained much higher accuracy of classification than commonly used manual methods.

However, for the practical application of presented artificial intelligence approaches by governmental units performing infrastructure inspections, the ability to train and deploy own models is crucial. From this point of view, Transfer Learning methods, obtaining only slightly lower accuracy than classic network training methods, offer a significant reduction in computing power demand during training and thus allow training of more classifiers in less time for various practical applications. Moreover, by utilizing robust feature extraction networks, proven on vast datasets like ImageNet, similar abilities to generalize knowledge will be obtained across various use cases. Thus, by utilizing KrakN framework, it is not necessary to fine-tune the networks hyperparameters each time the classifier is trained or utilize the knowledge specific to Machine Learning, that Bridge Inspectors are not expected not have --- this gives them the ability to train their own classifiers according to their needs.

However, it should be remembered that results presented above concern the worst case scenario of homogeneous training set. In order to obtain the highest network classification parameters, the dataset used for its training should consist of more inspection cases and thus, in practical application KrakN framework will most likely score even better metrics.

\subsection{Comparison with state of the art classifiers}

To assess the performance of the KrakN framework, we compared it to two state-of-the-art approaches -- CrackNet \cite{Zhang2018} and DeepCrack \cite{Liu2019} networks for detecting concrete cracks. It is however important to notice, that these networks are limited to crack detecting, while KrakN is an open framework, capable of detecting multiple types of defects. 

For the sake of comparison, each network without further retraining, was evaluated on two datasets. First one was the original KrakN evaluation dataset, featuring thin cracks of under 0.2 mm. The second was selected from datasets featured on CrackNet Github repository \cite{CrackNetdata}, and contained mostly of cracks wider than in the dataset delivered with KrakN. CrackNet dataset is composed of 40 thousand images. The goal of the performed experiment was to evaluate the state of the art approaches using different datasets and assess their ability to generalize knowledge in crack detecting. The results of the comparison are presented in Table~\ref{tab:comparative-performance-comp}.

\begin{table}[h!]
\begin{center}
\begin{tabular}{c c c c}
     & KrakN dataset [A/R] & CrackNet dataset [A/R] & Metric loss \\
     \hline
     \hline
     \multicolumn{1}{l}{KrakN} & 0.98 / 0.97 & 0.81 / 0.81 & \textbf{0.17 / 0.16} \\
     \hline
     \multicolumn{1}{l}{CrackNet} & 0.48 / 0.14 & 0.98 / 0.98 & 0.50 / 0.84 \\
     \hline
     \multicolumn{1}{l}{DeepCrack} & 0.26 / 0.40 & 0.97 / 0.79 & 0.71 / 0.39 \\
     \multicolumn{2}{l}{A -- Accuracy, R -- Recall}
\end{tabular}
\end{center}
\caption{Cross-dataset performance comparison.}
\label{tab:comparative-performance-comp}
\end{table}

As seen in Tab.~\ref{tab:comparative-performance-comp}, there is significant loss in performance when evaluating networks on unseen datasets. This is due to different abilities to generalize knowledge between seemingly identical tasks, as the width of cracks differ between the datasets. However, thanks to extracting more general features with the Transfer Learning approach, KrakN framework manages to achieve better performance when evaluated on data differing from its initial dataset. It is also possible thanks to the method of balancing weights with confidence threshold, that allows for seamlessly switching between sizes of the sought defect.

Additionally, when trained on CrackNet database, KrakN framework achieved value of \textbf{0.99} for both accuracy and recall when evaluated on CrackNet database subset. Then, without retraining it achieved \textbf{0.77} and \textbf{0.91} for accuracy and recall respectively when evaluated on KrakN dataset. It means that once again, metrics loss for KrakN framework which was equal to \textbf{(0.22 / 0.08)}, was much lower than for the other approaches.

It is also important to note a major limitation of current state-of-the-art classifiers that revealed itself during the tests. They were unable to process large scale images (over 120-megapixel) that were easily handled by KrakN, without exceeding 25 GB of VRAM memory --- an amount that could be found more commonly in specialized workstations for video processing and rendering. For this reason, before annotation, the texture had to be divided to up to 600 of image crops and then stitched back together after defect recognition. It was also found out that the computations were extremely resource heavy, as DeepCrack framework required a powerful GPU unit by default. Those however are scarce in governmental infrastructure management departments.

\subsection{KrakN in multi-stage classifier approach}

\begin{figure*}
    \begin{center}
        \includegraphics[width=\textwidth]{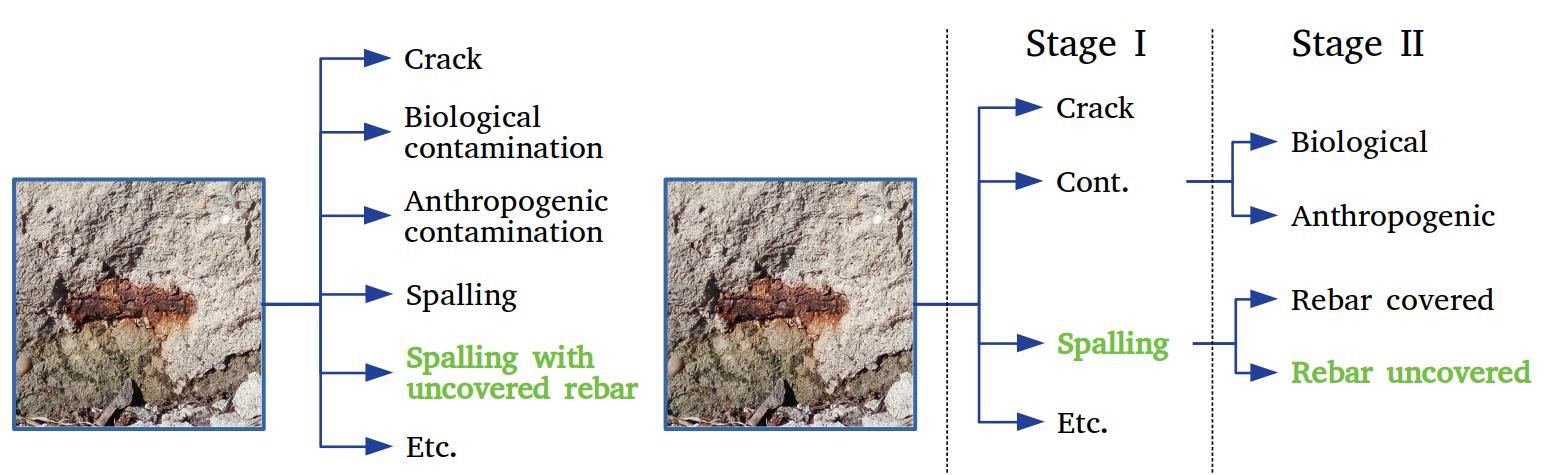}
    \end{center}
    \caption{Single-stage (left) classifier and an example of multi-stage classifier (right) reproducing the single-stage approach.}
    \label{fig:comparative-stages}
\end{figure*}

As the last part of comparative study, we attempted to demonstrate the versatility of the KrakN software package by reproducing the multi-classifier approach presented in~\cite{Huthwohl2019}. The main principle of multi-classifier approach is that by using multi-stage classifiers -- as compared to single-stage -- a higher accuracy can be obtained. Both of the approaches are shown in Fig.~\ref{fig:comparative-stages}

As seen in Fig.~\ref{fig:comparative-stages}, the multi-stage approach allows for more precise division between defect cases by grouping them into subsets for easier defects management. Furthermore, by training multiple classifiers, smaller datasets can be used, thus reducing training time for each classifier and in the case of adding new defect class to be recognized, only single classifier from the subset stage has to be retrained.

In order to employ multi-classification approach using KrakN, \code{network_trainer_Staged} directory in project repository~\cite{KrakN} provides necessary template. This implementation of KrakN allows using output images and masks from the initial classification to perform the next stage of image analysis utilizing different classifier to divide defects into subsets. 

For employing \code{network_train_Staged}, the user has to train new stage classifiers using the methods described in Section~\ref{sec:dataset-building} with a new dataset. Note that each of the classifiers in the subsequent stages has to be a separate model. For example, the second stage of image analysis depicted in Fig.~\ref{fig:comparative-stages} would consist of two classifiers -- separately for spalling and for contamination. Next, the classifier, in order for the software to detect it automatically, has to be renamed to match the defect name from the previous stage and placed in the \code{Classifiers} subdirectory of \code{network_trainer_Staged} directory. For the second stage of classification, the user has to place images, and their masks in \code{network_trainer_Staged} directory to the corresponding folders and run \code{moving_window_staged.py} script. As a result, new classification output consisting of classified images and masks will be created in \code{Output} directory. The \code{moving_window_staged.py} script will accept any number of images, masks, and classifiers.

As seen in the description above, KrakN framework proves to be versatile and susceptible to modifications that allow it to complete various tasks with only minor changes, while matching the functionality and remaining less computationally expensive than other approaches.

\section{Conclusions}\label{sec:conclusions}

The article describes main principles regarding the construction of datasets for artificial intelligence algorithms as well as the hands-on basis for using Transfer Learning techniques for training robust neural network algorithms. The work introduced an end to end approach to the defect detection and classification problem using Transfer Learned convolutional neural networks in the field of infrastructure inspections. The presented framework covers practical dataset construction, algorithm training and network implementation. Freely available software package KrakN, accompanying the paper, provides all of the described tools necessary for implementing the described methods by governmental units.

The article also compares the developed package to the network trained from scratch in the terms of both accuracy and the demand for computing power. This comparison demonstrated that similar ability to learn and generalize knowledge across different datasets can be obtained using both approaches. However, KrakN proves to be a much more viable solution considering the advantages of lower computing power demand and over 20-fold shorter calculation times. We also demonstrated that the developed software offers greater versatility than other state-of-the-art approaches as it achieved lower metrics loss across datasets and can be used for building multi-stage classifiers for further improving inspection accuracy while not being as resource-heavy. 


By making KrakN available for public use with the GitHub repository, its methods can be implemented by infrastructure management units at any infrastructure management level, regardless of its financial capabilities. With the use of artificial intelligence, the accuracy of assessments given to infrastructure members by inspectors can be significantly improved and unified across the infrastructure management units and inspectors, contributing to improving the infrastructure operational safety.

Public access to the developed algorithms may also contribute to increasing the popularity of the methods utilizing artificial intelligence in the field of Civil Engineering and thus allowing local governmental units to develop more efficient Structure Health Monitoring Systems based on image recognition and object classification. Furthermore, thanks to the methods being versatile and not limited to the single case of defects, it can cover the cases specific to each of the units common working scenarios.


Possible further development of the described software package will be concentrated on implementing it onto one of the mobile platforms. Thus it would be beneficial for further deployment of the software and can be used to accelerate the accumulation of data on infrastructure facilities. It can then be used for the research of more precise infrastructure deterioration models and ultimately lead to building fully autonomous, end to end SHM systems, that will be both accurate and affordable.

\paragraph{Acknowledgements}
This project was supported by Polish National Center for Research and Development under grant number POWR.03.05.00-00.z098/17-00 \textit{Silesian University of Technology as a Centre of Modern Education}. MZ and JM would like to thank to Ryszard Winiarczyk and Mateusz Ostaszewski for interesting discussions. Authors would like to thank Izabela Miszczak for proof reading the manuscript.

\bibliographystyle{ieeetr}
\bibliography{Bibliografia}

\end{document}